\documentclass[conference]{IEEEtran}

\usepackage{tikz}
\usepackage{times}
\usepackage{epsfig}
\usepackage{graphicx}
\usepackage{amsmath}
\usepackage{amssymb}
\usepackage{amsfonts}
\usepackage{multirow}
\usepackage{comment}
\usepackage{capt-of}
\usepackage{subcaption}
\usepackage{booktabs}
\usepackage{array}
\usepackage[font=small,labelfont=bf,tableposition=top]{caption}
\usepackage{stfloats}
\usepackage[normalem]{ulem}

\PassOptionsToPackage{hyphens}{url}\usepackage{hyperref}

\usetikzlibrary{positioning}

\definecolor{purp}{RGB}{112, 48, 160}

\IEEEoverridecommandlockouts

\begin{document}

\title{Learning to Segment Breast Biopsy Whole Slide Images}

\author{\IEEEauthorblockN{Sachin Mehta\IEEEauthorrefmark{1}\IEEEauthorrefmark{2},
Ezgi Mercan\IEEEauthorrefmark{1}\IEEEauthorrefmark{2}, Jamen Bartlett\IEEEauthorrefmark{3}, Donald Weaver\IEEEauthorrefmark{3}, Joann Elmore\IEEEauthorrefmark{2}, and Linda Shapiro\IEEEauthorrefmark{2}}
\IEEEauthorblockA{\IEEEauthorrefmark{2}University of Washington, Seattle, USA \\
Email: {\tt sacmehta@uw.edu, \{ezgi, shapiro\}@cs.washington.edu, jelmore@uw.edu}
}
\IEEEauthorblockA{\IEEEauthorrefmark{3}University of Vermont, Burlington, VT \\
Email: {\tt \{jamen.bartlett,donald.weaver\}@uvmhealth.org}
}
\IEEEauthorblockA{\thanks{\IEEEauthorrefmark{1}Authors contributed equally.}}
}

\def\etal{\emph{et al.}}

\maketitle

\begin{abstract}
We trained and applied an encoder-decoder model to semantically segment breast biopsy images into biologically meaningful tissue labels. Since conventional encoder-decoder networks cannot be applied directly on large biopsy images and the different sized structures in biopsies present novel challenges, we propose four modifications: (1) an input-aware encoding block to compensate for information loss, (2) a new dense connection pattern between encoder and decoder, (3) dense and sparse decoders to combine multi-level features, (4) a multi-resolution network that fuses the results of encoder-decoders run on different resolutions. Our model outperforms a feature-based approach and conventional encoder-decoders from the literature. We use semantic segmentations produced with our model in an automated diagnosis task and obtain higher accuracies than a baseline approach that employs an SVM for feature-based segmentation, both using the same segmentation-based diagnostic features.
\end{abstract}

\section{Introduction}
Breast cancer is traditionally diagnosed with  histopathological interpretation of the biopsy samples on glass slides by pathologists. Whole slide imaging (WSI) is a technology that captures the contents of glass slides in a multi-resolution image. With the developments in  whole slide imaging, it is now possible to develop computer-aided diagnostic tools that support the decision-making process of medical experts. Until recently, the use of WSIs was limited to non-clinical purposes such as research, education, obtaining second opinions, and archiving, but they have been approved for diagnostic use in the US starting April 2017 \cite{website:fda}. 

\begin{figure}[!t]
  \centering
  \begin{tabular}{c}
  \includegraphics[width=0.8\columnwidth]{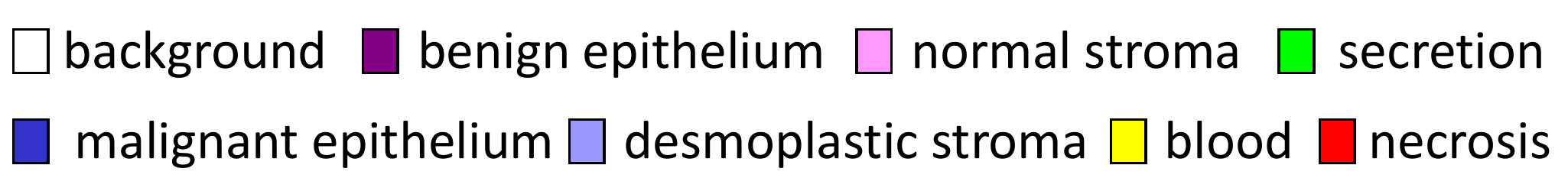}\\
   \includegraphics[width=0.8\columnwidth]{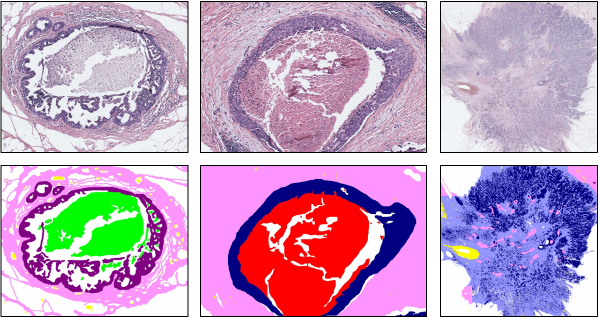}
   \end{tabular}
  \caption{The set of tissue labels used in semantic segmentation: (top row) three example cases from the dataset and (bottom row) the pixel labels provided by a pathologist. Best viewed in color.}
  \label{fig:tissue_labels}
\end{figure}

Automated cancer detection from digital slides is a well-studied task in the computer vision community \cite{CruzRoa2017AccurateAR} and several image datasets have been developed for malignant tumors \cite{website:tcga, VETA2015237, website:tupac}; however, little work exists in differentiating the full spectrum of breast lesions from benign to pre-invasive lesions, and to invasive cancer \cite{dong2014computational}. Pre-invasive lesions presents a more difficult classification scenario than the binary classification task of invasive cancer detection. It requires careful analysis of epithelial structures in the breast biopsy images. In this paper, we propose a state-of-the-art semantic segmentation system to produce a tissue label image (Figure \ref{fig:tissue_labels}) for the WSIs of breast biopsies that can lead to an automated diagnosis system. 

Our system builds on the encoder-decoder networks that are the state-of-the-art approaches for semantic segmentation. However, conventional architectures are not directly applicable to whole slide breast biopsy images with dimensions in gigapixels. A sliding window approach to crop fixed-sized images from WSIs is promising \cite{hou2016patch:eke}, but dividing the large structures limits the context available to CNNs and affect the segmentation performance. Unlike general image datasets (e.g. \cite{krizhevsky2012imagenet,Everingham2015, Sirinukunwattana16:eke}), breast biopsy images have objects of interest in varied sizes. For some WSIs, the diagnosis is made while looking at the whole image, while others require the detection of a small structure at high resolutions. Simply using a sliding window with a constant size causes loss of information available at different resolutions. 

This paper proposes a new multi-resolution encoder-decoder architecture that was specifically designed to handle the challenges of the breast biopsy semantic segmentation problem. The architecture is described in detail, and a rigorous set of experiments is applied to compare its segmentation performance to multiple different other models. Finally, the network is used in a set of diagnostic classification experiments that further show its benefits.

\section{Related Work}
Following the success of CNNs in image classification tasks \cite{szegedy2015going:eke, simonyan2014very:eke, he2016deep:eke}, they have been extended for dense prediction tasks such as semantic segmentation \cite{shelhamer2017fully:eke, noh2015learning:eke, Badrinarayanan2017:eke}. Unlike object proposal-based methods \cite{girshick2014rich, hariharan2015hypercolumns}, fully convolutional networks (FCN) have enabled end-to-end training and have shown efficient feature learning. These methods are widely used for segmenting both natural \cite{shelhamer2017fully:eke, noh2015learning:eke, Badrinarayanan2017:eke, chen2016deeplab} and medical images \cite{Ronneberger2015:eke, fakhry2017residual:eke, milletari2016v, yu2017volumetric}. 

FCN-based networks generate coarse segmentation masks and several techniques have been proposed to address this limitation such as skip-connections \cite{shelhamer2017fully:eke, Ronneberger2015:eke, fakhry2017residual:eke}, atrous/dilated convolutions \cite{chen2016deeplab, YuKoltun2016:eke}, deconvolutional networks \cite{noh2015learning:eke, Badrinarayanan2017:eke, fakhry2017residual:eke, Ronneberger2015:eke, Chen2016:eke}, and multiple input networks (e.g. different scales \cite{ Chen_2016_CVPR, Zagoruyko2016Multipath, lin2016refinenet} or streams \cite{farabet2013learning}). These methods process the input sources either independently \cite{chen2016deeplab, Chen_2016_CVPR, farabet2013learning, karpathy2014large} or recursively \cite{pinheiro2014recurrent, eigen2015predicting}; thus exploit the features from multiple levels to refine the segmentation masks. Additionally, conditional random fields (CRFs) have been used to further refine the segmentation results \cite{zheng2015conditional, chen2016deeplab, YuKoltun2016:eke}.

Several CNN-based methods have been applied for segmenting medical images (e.g. EM \cite{Ronneberger2015:eke}, brain \cite{fakhry2017residual:eke}, gland \cite{Chen2016:eke}, and 3D MR  \cite{yu2017volumetric} images). Yet, segmenting breast biopsy images, with a full range of diagnosis from benign to invasive, still remains a challenge. Our approach applies previous work on encoder-decoders (e.g. \cite{Badrinarayanan2017:eke, fakhry2017residual:eke}) and improves upon them with carefully designed components that address their limitations on WSI applications. 

\section{Breast Biopsy Dataset}
\label{sec:dataset}
Our dataset contains 240 breast biopsies selected from the Breast Cancer Surveillance Consortium \cite{BCSC} archives in New Hampshire and Vermont. The cases span a wide range of diagnoses that mapped to four diagnostic categories: benign, atypia, ductal carcinoma \textit{in-situ} (DCIS), and invasive cancer. The original H\&E (heamatoxylin and eosin) stained glass slides were scanned using an iScan CoreoAu\textsuperscript{\textregistered} in $40\times$ magnification. A technician and an experienced breast pathologist reviewed each digital image, rescanning as needed to obtain the highest quality. The average image size for the 240 WSIs was $90,000 \times 70,000$ pixels.

All 240 digital slides were interpreted by an expert panel of three pathologists to produce an expert consensus diagnosis for each case. Experts also provided one or more regions of interest (ROIs) supporting the expert consensus diagnosis on each WSI. Since some cases had more than one ROI per WSI, the final set includes 102 benign, 128 atypia, 162 DCIS and 36 invasive ROIs.

\begin{table}[b!]
\centering
\small
\begin{tabular}{rcccc}
    \hline
    \textbf{Diagnostic }& \textbf{\#ROI} & \textbf{\#ROI} & \textbf{\#ROI} & \textbf{Avg. size}\\ 
    \textbf{Category}&     (training)         & (test)      & (total)    &  (pixels) \\
    \midrule
    Benign   & 4  & 5  & 9  & $9K \times 9K$   \\
    Atypia   & 11 & 11 & 22 & $6K \times 7K$   \\ 
    DCIS     & 12 & 10 & 22 & $8K \times 10K$  \\
    Invasive &  3 &  2 &  5 & $38K \times 44K$ \\ 
    \hline
    \textbf{Total} & \textbf{30} & \textbf{28} & \textbf{58} & \textbf{$10K \times 12K$} \\
    \hline
\end{tabular}
\caption{Distribution of diagnostic categories and average image sizes from the segmentation subset.}
\label{tab:datastats}
\end{table}

To describe the structural changes that lead to cancer in the breast tissue, we produced a set of eight tissue labels in collaboration with an expert pathologist: 
(1) \textit{benign epithelium}: the epithelial cells in the benign and atypia categories, (2) \textit{malignant epithelium}: the bigger and more irregular epithelial cells from the DCIS and invasive cancer categories, (3) \textit{normal stroma}: the connective tissue between the regular ductal structures in the breast, (4) \textit{desmoplastic stroma}: proliferated stromal cells associated with tumor, (5) \textit{secretion}: benign substance secreted from the ducts, (6) \textit{necrosis} the dead cells at the center of the ducts in the DCIS and invasive cases, (7) \textit{blood}: the blood cells, which are rare but have a very distinct appearance, and (8) \textit{background}: the pixels that do not contain any tissue.

Although some labels are not critical for diagnosis, our tissue label set was intended to cover all the pixels in the images. Due to the expertise needed for labeling and the size of the biopsy images, we randomly selected a subset of 40 cases (58 ROIs) to be annotated by a pathologist. Table \ref{tab:datastats} summarizes the distribution of four diagnostic categories in training and test sets as well as average image sizes. Figure \ref{fig:tissue_labels} shows three example images along with their pixel-wise labels provided by the pathologist.

\section{Background}
\label{sec:background}
Encoder-decoder networks are state-of-the-art networks for segmenting 2D (e.g. \cite{Ronneberger2015:eke, fakhry2017residual:eke}) as well as 3D (e.g. \cite{milletari2016v, yu2017volumetric}) medical images.
In a conventional encoder, the transition between two subsequent encoding blocks, $l^{th}$ and $(l+1)^{th}$, can be formulated as \cite{krizhevsky2012imagenet, simonyan2014very:eke}: $\mathbf{x}^{l+1}_e = \mathcal{F}_e(\mathbf{x}^l_e)$. 
In a class of encoder networks, called residual networks, the input and output of the $l^{th}$ block are combined to improve the gradient flow \cite{he2016deep:eke}: 
\begin{equation}
\mathbf{x}^{l+1}_e = \mathcal{F}_e(\mathbf{x}^l_e) + \mathbf{x}^l_e
\label{eq:rcu}
\end{equation}
where $\mathcal{F}_e(\mathbf{x}^l_e)$ is a function comprising two $3 \times 3$ convolution operations. This block is referred as a Residual Convolutional Unit (RCU) (Figure \ref{fig:rcu}). 

\begin{figure}[b!]
\centering
\begin{subfigure}[b]{0.48\columnwidth}
\centering
\includegraphics[height=100px]{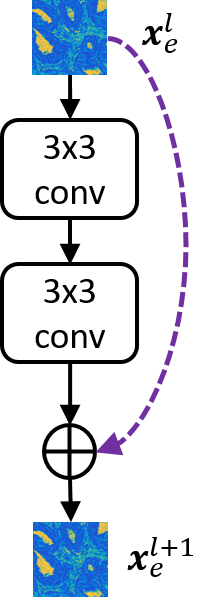}
\caption{\small{RCU} \cite{he2016deep:eke}}
\label{fig:rcu}
\end{subfigure}  
\hfill
\begin{subfigure}[b]{0.48\columnwidth}
\centering
\includegraphics[height=100px]{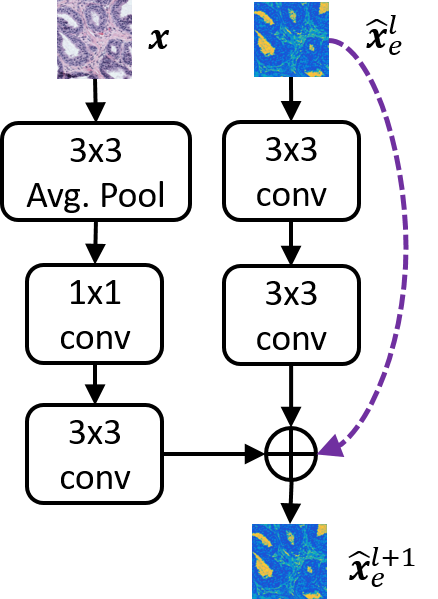}
\caption{\small{Our Input-Aware RCU}}
\label{fig:iarcu}
\end{subfigure}
\caption{Different type of encoding blocks: (a) residual convolutional unit (RCU) and (b) the proposed input-aware residual convolutional unit (IA-RCU). }
\end{figure}
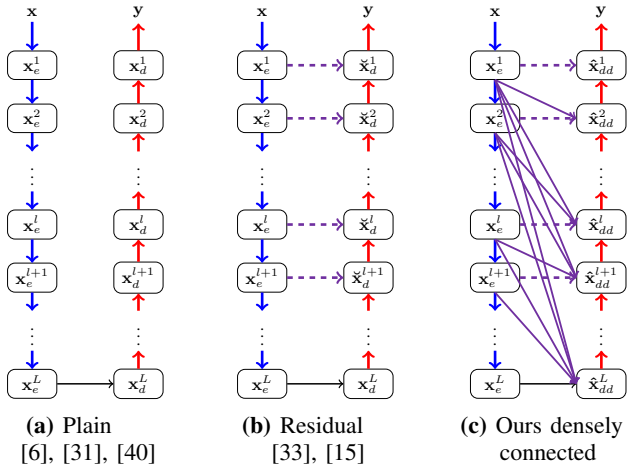
\begin{figure}[t!]
\small{
\centering
 \begin{subfigure}[b]{0.3\columnwidth}
\centering
\resizebox{!}{150px}{\begin{tikzpicture}[
block/.style={
text width={width("Temp")},
align=center,rounded corners,
font=\small}]
\draw (0,6) node[block] (x) {$\mathbf{x}$};
\draw (0,5) node[block,draw, fill=white] (x1) {$\mathbf{x}_e^1$};
\draw (0,4) node[block,draw, fill=white] (x2) {$\mathbf{x}_e^2$};
\draw (0,3) node[block] (x3) {$\vdots$};
\draw (0,2) node[block,draw, fill=white] (x4) {$\mathbf{x}_e^l$};
\draw (0,1) node[block,draw, fill=white] (x5) {$\mathbf{x}_e^{l+1}$};
\draw (0,0) node[block] (x6) {$\vdots$};
\draw (0,-1) node[block,draw, fill=white] (x7) {$\mathbf{x}_e^{L}$};

\draw[thick, ->,color=blue, line width=0.5mm] (x) -- (x1);
\draw[thick, ->, color=blue, line width=0.5mm] (x1) -- (x2);
\draw[thick, ->, color=blue, line width=0.5mm] (x2) -- (x3);
\draw[thick, ->, color=blue, line width=0.5mm] (x3) -- (x4);
\draw[thick, ->, color=blue, line width=0.5mm] (x4) -- (x5);
\draw[thick, ->, color=blue, line width=0.5mm] (x5) -- (x6);
\draw[thick, ->, color=blue, line width=0.5mm] (x6) -- (x7);

\draw (2,6) node[block] (x0) {$\mathbf{y}$};
\draw (2,5) node[block,draw, fill=white] (x01) {$\mathbf{x}_d^1$};
\draw (2,4) node[block,draw, fill=white] (x02) {$\mathbf{x}_d^2$};
\draw (2,3) node[block] (x03) {$\vdots$};
\draw (2,2) node[block,draw, fill=white] (x04) {$\mathbf{x}_d^l$};
\draw (2,1) node[block,draw, fill=white] (x05) {$\mathbf{x}_d^{l+1}$};
\draw (2,0) node[block] (x06) {$\vdots$};
\draw (2,-1) node[block,draw, fill=white] (x07) {$\mathbf{x}_d^{L}$};

\draw[thick, ->, color=red, line width=0.5mm] (x07) -- (x06);
\draw[thick, ->, color=red, line width=0.5mm] (x06) -- (x05);
\draw[thick, ->, color=red, line width=0.5mm] (x05) -- (x04);
\draw[thick, ->, color=red, line width=0.5mm] (x04) -- (x03);
\draw[thick, ->, color=red, line width=0.5mm] (x03) -- (x02);
\draw[thick, ->, color=red, line width=0.5mm] (x02) -- (x01);
\draw[thick, ->, color=red, line width=0.5mm] (x01) -- (x0);
\draw[thick, ->] (x7) -- (x07);

\end{tikzpicture}} 
\caption{\centering {Plain \newline \cite{Badrinarayanan2017:eke, noh2015learning:eke, yang2016object:eke}} }
\label{fig:plainEncDec}
\end{subfigure}
\hfill
\begin{subfigure}[b]{0.32\columnwidth}
\centering
\resizebox{!}{150px}{\begin{tikzpicture}[
block/.style={
text width={width("Temp")},
align=center,rounded corners,
font=\small}]
\draw (0,6) node[block] (x) {$\mathbf{x}$};
\draw (0,5) node[block,draw, fill=white] (x1) {$\mathbf{x}_e^1$};
\draw (0,4) node[block,draw, fill=white] (x2) {$\mathbf{x}_e^2$};
\draw (0,3) node[block] (x3) {$\vdots$};
\draw (0,2) node[block,draw, fill=white] (x4) {$\mathbf{x}_e^l$};
\draw (0,1) node[block,draw, fill=white] (x5) {$\mathbf{x}_e^{l+1}$};
\draw (0,0) node[block] (x6) {$\vdots$};
\draw (0,-1) node[block,draw, fill=white] (x7) {$\mathbf{x}_e^{L}$};

\draw[thick, ->,color=blue, line width=0.5mm] (x) -- (x1);
\draw[thick, ->, color=blue, line width=0.5mm] (x1) -- (x2);
\draw[thick, ->, color=blue, line width=0.5mm] (x2) -- (x3);
\draw[thick, ->, color=blue, line width=0.5mm] (x3) -- (x4);
\draw[thick, ->, color=blue, line width=0.5mm] (x4) -- (x5);
\draw[thick, ->, color=blue, line width=0.5mm] (x5) -- (x6);
\draw[thick, ->, color=blue, line width=0.5mm] (x6) -- (x7);

\draw (2,6) node[block] (x0) {$\mathbf{y}$};
\draw (2,5) node[block,draw, fill=white] (x01) {$\mathbf{\breve{x}}_d^1$};
\draw (2,4) node[block,draw, fill=white] (x02) {$\mathbf{\breve{x}}_d^2$};
\draw (2,3) node[block] (x03) {$\vdots$};
\draw (2,2) node[block,draw, fill=white] (x04) {$\mathbf{\breve{x}}_d^l$};
\draw (2,1) node[block,draw, fill=white] (x05) {$\mathbf{\breve{x}}_d^{l+1}$};
\draw (2,0) node[block] (x06) {$\vdots$};
\draw (2,-1) node[block,draw, fill=white] (x07) {$\mathbf{x}_d^{L}$};

\draw[thick, ->, color=red, line width=0.5mm] (x07) -- (x06);
\draw[thick, ->, color=red, line width=0.5mm] (x06) -- (x05);
\draw[thick, ->, color=red, line width=0.5mm] (x05) -- (x04);
\draw[thick, ->, color=red, line width=0.5mm] (x04) -- (x03);
\draw[thick, ->, color=red, line width=0.5mm] (x03) -- (x02);
\draw[thick, ->, color=red, line width=0.5mm] (x02) -- (x01);
\draw[thick, ->, color=red, line width=0.5mm] (x01) -- (x0);
\draw[thick, ->] (x7) -- (x07);

\draw[dashed, ->, color=purp, line width=0.5mm] (x5) -- (x05);
\draw[dashed, ->, color=purp, line width=0.5mm] (x4) -- (x04);
\draw[dashed, ->, color=purp, line width=0.5mm] (x2) -- (x02);
\draw[dashed, ->, color=purp, line width=0.5mm] (x1) -- (x01);

\end{tikzpicture}}
\caption{\centering{Residual \newline \cite{Ronneberger2015:eke, fakhry2017residual:eke} }}
\label{fig:resEncDec}
\end{subfigure}  
\hfill
\begin{subfigure}[b]{0.3\columnwidth}
\centering
\resizebox{!}{150px}{\definecolor{purp}{RGB}{112, 48, 160}
\begin{tikzpicture}[
block/.style={
text width={width("Temp")},
align=center,rounded corners,
font=\small}]
\draw (0,6) node[block] (x) {$\mathbf{x}$};
\draw (0,5) node[block,draw, fill=white] (x1) {$\mathbf{x}_e^1$};
\draw (0,4) node[block,draw, fill=white] (x2) {$\mathbf{x}_e^2$};
\draw (0,3) node[block] (x3) {$\vdots$};
\draw (0,2) node[block,draw, fill=white] (x4) {$\mathbf{x}_e^l$};
\draw (0,1) node[block,draw, fill=white] (x5) {$\mathbf{x}_e^{l+1}$};
\draw (0,0) node[block] (x6) {$\vdots$};
\draw (0,-1) node[block,draw, fill=white] (x7) {$\mathbf{x}_e^{L}$};

\draw[thick, ->,color=blue, line width=0.5mm] (x) -- (x1);
\draw[thick, ->, color=blue, line width=0.5mm] (x1) -- (x2);
\draw[thick, ->, color=blue, line width=0.5mm] (x2) -- (x3);
\draw[thick, ->, color=blue, line width=0.5mm] (x3) -- (x4);
\draw[thick, ->, color=blue, line width=0.5mm] (x4) -- (x5);
\draw[thick, ->, color=blue, line width=0.5mm] (x5) -- (x6);
\draw[thick, ->, color=blue, line width=0.5mm] (x6) -- (x7);

\draw (2,6) node[block] (x0) {$\mathbf{y}$};
\draw (2,5) node[block,draw, fill=white] (x01) {$\mathbf{\hat{x}}_{dd}^1$};
\draw (2,4) node[block,draw, fill=white] (x02) {$\mathbf{\hat{x}}_{dd}^2$};
\draw (2,3) node[block] (x03) {$\vdots$};
\draw (2,2) node[block,draw, fill=white] (x04) {$\mathbf{\hat{x}}_{dd}^l$};
\draw (2,1) node[block,draw, fill=white] (x05) {$\mathbf{\hat{x}}_{dd}^{l+1}$};
\draw (2,0) node[block] (x06) {$\vdots$};
\draw (2,-1) node[block,draw, fill=white] (x07) {$\mathbf{\hat{x}}_{dd}^{L}$};

\draw[thick, ->, color=red, line width=0.5mm] (x07) -- (x06);
\draw[thick, ->, color=red, line width=0.5mm] (x06) -- (x05);
\draw[thick, ->, color=red, line width=0.5mm] (x05) -- (x04);
\draw[thick, ->, color=red, line width=0.5mm] (x04) -- (x03);
\draw[thick, ->, color=red, line width=0.5mm] (x03) -- (x02);
\draw[thick, ->, color=red, line width=0.5mm] (x02) -- (x01);
\draw[thick, ->, color=red, line width=0.5mm] (x01) -- (x0);
\draw[thick, ->] (x7) -- (x07);

\draw[thick, ->, color=purp, line width=0.35mm] (x1.south) -- (x07.west);
\draw[thick, ->, color=purp, line width=0.35mm] (x1.south) -- (x05.west);
\draw[thick, ->, color=purp, line width=0.35mm] (x1.south) -- (x04.west);
\draw[thick, ->, color=purp, line width=0.35mm] (x1.south) -- (x02.west);

\draw[thick, ->, color=purp, line width=0.35mm] (x2.south) -- (x07.west);
\draw[thick, ->, color=purp, line width=0.35mm] (x2.south) -- (x05.west);
\draw[thick, ->, color=purp, line width=0.35mm] (x2.south) -- (x04.west);

\draw[thick, ->, color=purp, line width=0.35mm] (x4.south) -- (x07.west);
\draw[thick, ->, color=purp, line width=0.35mm] (x4.south) -- (x05.west);

\draw[thick, ->, color=purp, line width=0.35mm] (x5.south) -- (x07.west);

\draw[dashed, ->, color=purp, line width=0.5mm] (x5) -- (x05);
\draw[dashed, ->, color=purp, line width=0.5mm] (x4) -- (x04);
\draw[dashed, ->, color=purp, line width=0.5mm] (x2) -- (x02);
\draw[dashed, ->, color=purp, line width=0.5mm] (x1) -- (x01);

\end{tikzpicture}}
\caption{\centering{Ours densely \newline connected}}
\label{fig:denseEncDec}
\end{subfigure}  
}
\caption{(a, b) Conventional and (c) ours densely connected encoder-decoder networks with $L$ encoding and decoding blocks. These networks take an input $\mathbf{x}$ and generates an output $\mathbf{y}$. Here, {\Large$\color{purp}{\dashrightarrow}$} and {\Large$\color{purp}{\rightarrow}$} represents  residual and dense links between the encoder and the decoder.}
\label{fig:tradEncDec}
\end{figure}

In a conventional decoder (Figure \ref{fig:plainEncDec}), the transition between two subsequent decoding blocks, $l^{th}$ and $(l+1)^{th}$, can be formulated as \cite{shelhamer2017fully:eke, chen2016deeplab,noh2015learning:eke,Badrinarayanan2017:eke}: $\mathbf{x}^{l}_d = \mathcal{F}_d(\mathbf{x}^{l+1}_d)$. To improve the gradient flow between the encoder and the decoder, the output of the $l^{th}$ encoding block and the corresponding decoding block can be combined as \cite{fakhry2017residual:eke, Ronneberger2015:eke}:
\begin{equation}
\mathbf{\breve{x}}^{l}_d = \mathbf{x}^{l}_e + \mathcal{F}_d(\mathbf{x}^{l+1}_d) 
\label{eq:resLink}
\end{equation}
where $\mathcal{F}_d(\mathbf{x}^l_d)$ is a decoding function that performs a $3 \times 3$ deconvolution operation. Such an encoder-decoder network with skip-connections between encoding and decoding blocks is called a \textit{residual encoder-decoder} (Figure \ref{fig:resEncDec}). 
The deconvolution operation: (1) up-samples the feature maps, and (2) reduces the dimensionality of the feature maps. Note that the deconvolutional filters are capable of learning the non-linear up-sampling operations \cite{shelhamer2017fully:eke}.

\begin{figure*}[t!]
\centering
\resizebox{1.6\columnwidth}{!}{
\includegraphics[width=\columnwidth]{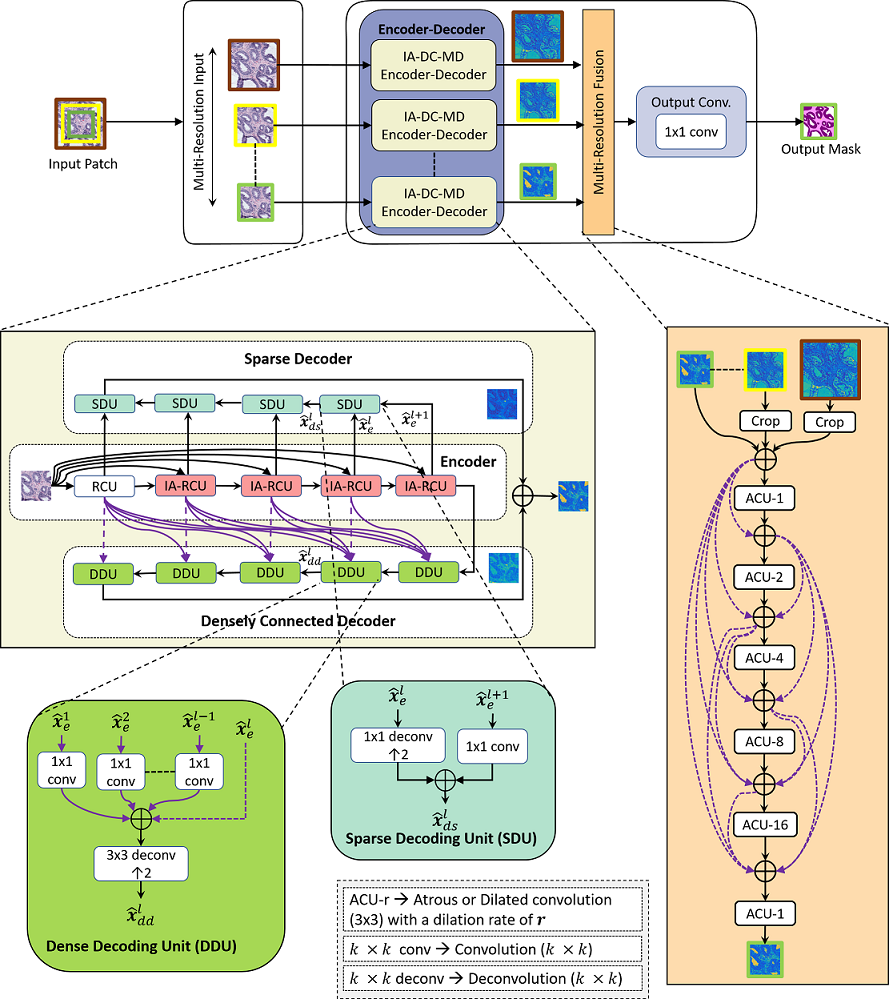}
}
\caption{Our multi-resolution encoder-decoder network that incorporates input-aware encoding blocks, sparse and densely connected decoding networks, and densely connected fusion block. Different components in our architecture makes use of identity and projection mappings; thereby helping in back-propagating the information directly to the input paths efficiently. {\Large$\color{purp}{\rightarrow}$} and  {\Large$\color{purp}{\dashrightarrow}$} links denotes the identity and projection links. The number of channels at different levels of encoder, densely connected decoder, and sparse decoder follow the following sequences:  $64\rightarrow64\rightarrow128\rightarrow256\rightarrow512$, $256\rightarrow128\rightarrow64\rightarrow64\rightarrow C$, and $C\rightarrow C\rightarrow C\rightarrow C$. Best viewed in color.}
\label{fig:proposedModel}
\end{figure*}

\section{Proposed Encoder-Decoder Network}
\label{sec:cnnSeg}
We propose a new encoder-decoder architecture to address the challenges that semantic segmentation of breast biopsies presents. Our network incorporates four new features: (1) input-aware encoding blocks (IA-RCU) that reinforces the input inside the encoder to compensate the loss of information due to down-sampling operations, (2) a densely connected decoding network and (3) an additional sparsely connected decoding network to efficiently combine the multi-level features aggregated by the encoder, and (4) a multi-resolution network for context-aware learning, which combines the output of different resolutions using a densely connected fusion block. Our network makes use of long-range skip-connections with identity and projection mappings in the encoder, the decoder, and fusion block to efficiently back-propagate the information to the input source and prevent the vanishing gradients; thereby helps train our network efficiently end-to-end. An overview of our network is illustrated in Figure \ref{fig:proposedModel} with details below.
\subsection{Input-aware encoding blocks (IA-RCU)}
The down-sampling operations in the encoder result in a loss of spatial information. To compensate the loss of spatial information, we introduce an input-aware encoding block (IA-RCU) that reinforces the input image at different levels of the encoder for better encoding of the spatial relationships and learned features. The IA-RCU, sketched in Figure \ref{fig:iarcu}, introduces an additional path which can be viewed as a different connectivity pattern that establishes a direct link between an input image and any encoding stage, making each encoding block \textit{aware of the input image}; thereby allowing gradients to flow back directly to the input paths. Additionally, the IA-RCU allows the encoding blocks to learn the features relevant to the input. The IA-RCU can be mathematically defined as:
\begin{equation}
 \mathbf{\hat{x}}_{e}^{l+1} =   \mathbf{{x}}_e^{l+1} + \mathcal{F}_{IA}(\mathbf{x})
\end{equation}
where $\mathcal{F}_{IA}(\mathbf{x})$ represents an input-aware mapping to be learned. $\mathcal{F}_{IA}$ is a composite function comprising a $3\times3$ average pooling operation that sub-samples the input image $\mathbf{x}$ to the same size as the encoding block $\mathbf{{x}}_e^{l+1}$, followed by $1\times1$ and $3\times3$ convolution operations that first projects the sub-sampled image to the same vector space as the encoding block $\mathbf{{x}}_e^{l+1}$ (Eq. \ref{eq:rcu}) and then computes the dense features. 
\subsection{Densely Connected Decoding Blocks}
Unlike a plain encoder-decoder network (Figure \ref{fig:plainEncDec}), the skip-connections in the residual encoder-decoder network (Figure \ref{fig:resEncDec}) establishes a direct link between the encoding block and corresponding decoding block, which helps to improve the information flow. To further improve the information flow, we introduce direct connections between a decoding block and all encoding blocks that are at the same or lower level (Figure \ref{fig:denseEncDec}). The $l^{th}$ decoding block receives the output feature maps from encoding blocks $1$ to $l$. Dense connections can be defined as a modification to Eq. \ref{eq:resLink}: 
\begin{equation}
\mathbf{\hat{x}}^{l}_{dd} = \mathcal{F}_d(\mathbf{x}^{l+1}_d) + \sum_{i=1}^{l} \mathcal{F}_D(\mathbf{\hat{x}}^{i}_e)
\label{eq:denseLink}
\end{equation}
$\mathcal{F}_D( \mathbf{\hat{x}}^{i}_e)$  is the dense connection mapping to be learned. $\mathcal{F}_D$ consists of a $1\times1$ convolution operation, which projects the feature maps of the $i^{th}$ encoding block $\mathbf{\hat{x}}^{i}_e$ to the same vector space as $\mathbf{x}^{l}_d$.

\subsection{Multiple Decoding Paths}
For a given input image $\mathbf{x}$, we aim to efficiently combine the low- and mid-level features of the encoding network with high-level features to generate a pixel-level semantic segmentation mask. To do so, we must invert the loss of resolution from down-sampling. Using previous work \cite{noh2015learning:eke, Badrinarayanan2017:eke, Ronneberger2015:eke, fakhry2017residual:eke}, we augment the encoder network with the bottom-up refinement approach. We introduce two decoding networks, densely connected and sparse, that decode the encoded input into a $C$-dimensional output, where $C$ represents the number of classes in the dataset. Figure \ref{fig:proposedModel} shows our network with multiple decoding paths. 

The densely connected decoder stacks the densely connected decoding blocks, defined in Eq. \ref{eq:denseLink}, to decode the encoded feature maps into $C$-dimensional space.  Because of the dense connections between the encoder and the decoder, we call this decoder a \textit{densely connected decoder}. 
The sparse decoder projects the high-dimensional feature maps of each encoding block into $C$-dimensional vector spaces, which are then combined using a bottom-up approach. A sparse decoding function $\mathcal{F}_S$ can be formulated as:
\begin{equation}
\mathbf{\hat{x}}_{ds}^{l} = \mathcal{F}_S(\{\mathbf{\hat{x}}_e^{l}, \mathbf{\hat{x}}_e^{l+1}\})
\label{eq:sparse}
\end{equation}
$\mathcal{F}_S(\{\mathbf{\hat{x}}_e^{l}, \mathbf{\hat{x}}_e^{l+1}\})$ is a function consisting of $1\times1$ deconvolutional and convolutional operations that projects high-dimensional encoder feature maps to $C$-dimensional vector space. Additionally, deconvolution operation up-samples the feature maps of $\mathbf{\hat{x}}_e^{l+1}$ to the same size as $\mathbf{\hat{x}}_e^{l}$. Because of the $1\times1$ convolution/deconvolutional operations involved, we call this decoder a \textit{sparse decoder}.

\subsection{Multiple Resolution Input}
A sliding-window approach is promising for segmenting large biopsy images, however, the size of the patch determines the context available to the CNN model. Such an approach divides the bigger structures into smaller patches and may hurt the performance of the CNN method, especially at the border of the patch. To make the CNN model aware of the surrounding information, we introduce a multi-resolution network, which consists of the composition of $P$ instances of the encoder-decoder network (Figure \ref{fig:proposedModel}). The $p^{th}$ instance takes the input patch $\mathbf{x}_p$ and generates the $C$-dimensional output $\mathbf{y}_p$. The spatial dimensions of each instance are different. A cropping function $\mathcal{F}_{Cr}(\mathbf{y}_p)$ takes the output of the $p^{th}$ instance and centrally crops it to produce the output $\mathbf{\hat{y}}_p$, which has the same dimensions as $\mathbf{y}_{P}$. After cropping, a multi-resolution fusion function $\mathcal{F}_{Mr}(\{\mathbf{\hat{y}}_1, \cdots, \mathbf{\hat{y}}_{P-1}, \mathbf{y}_P \})$ is applied to fuse the output of these $P$ network instances to produce the output $\mathbf{y}^{pred}$.

The multi-resolution fusion function $\mathcal{F}_{Mr}$, visualized in Figure \ref{fig:proposedModel}, first combines the $P$ instances using an element-wise sum operation and then extracts the dense features using a stack of $3\times3$ dilated or atrous convolution operations with different dilation rates $r$. A traditional context module \cite{YuKoltun2016:eke} may suffer from degradation problem and impede the information flow. Following Huang \etal  \cite{huang2017densely}, we introduce direct identity mappings from any layer to its subsequent layers to improve the information flow in the fusion block. We combine the output of any layer with the preceding layers using an element-wise sum operation.

\section{Experiments and Results}
\begin{figure}[b!]
\centering
\begin{subfigure}[b]{0.24\columnwidth}
\centering
\includegraphics[height=150px]{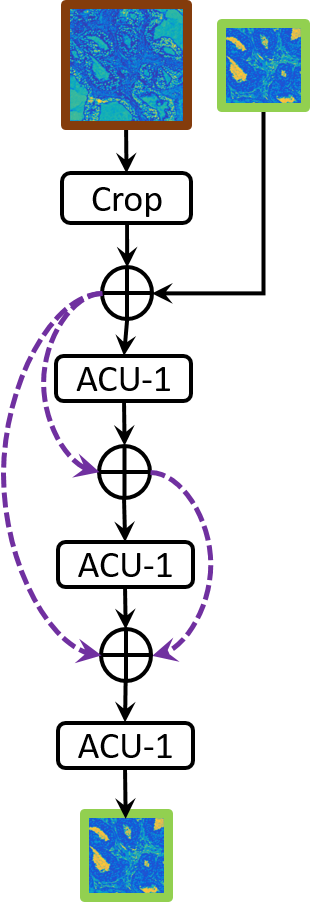}
\caption{Fusion-A}
\label{fig:f1}
\end{subfigure}
\hfill
\begin{subfigure}[b]{0.72\columnwidth}
\centering
\includegraphics[height=150px]{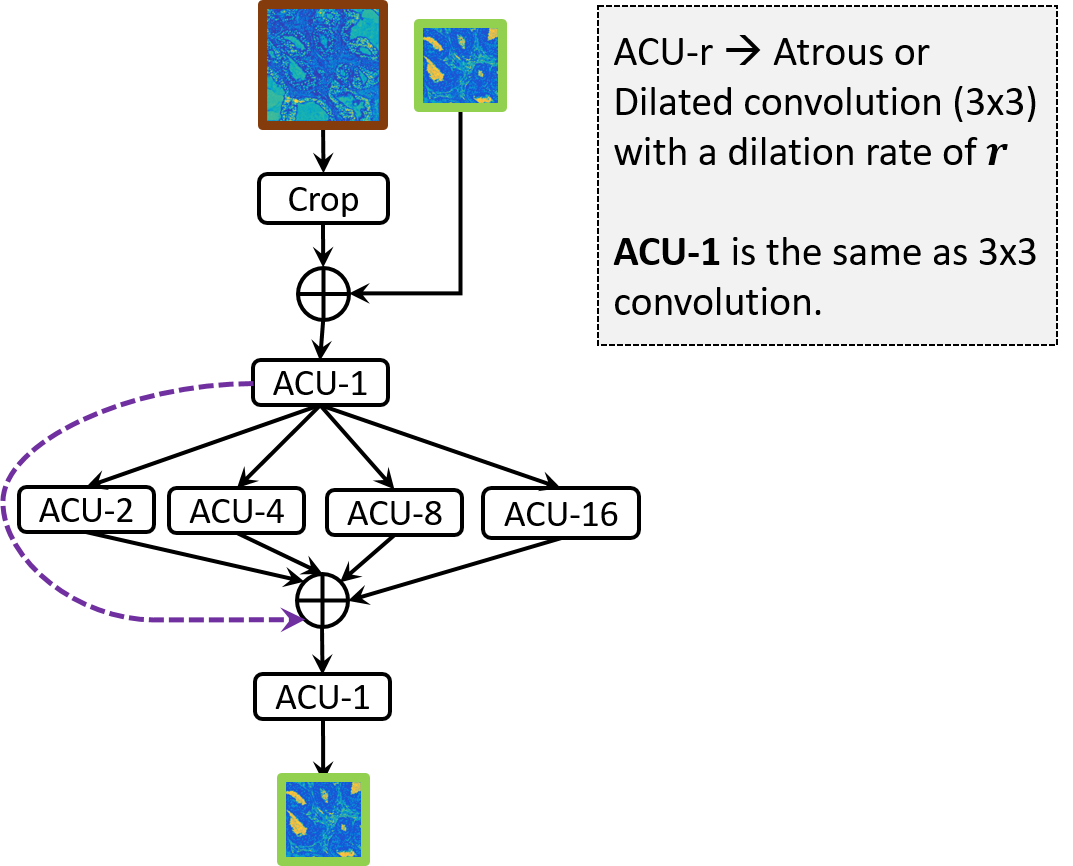}
\caption{Fusion-B}
\label{fig:f2}
\end{subfigure}
\caption{Different fusion strategies for multi-resolution network.}
\label{fig:fuseTech}
\end{figure}

To evaluate each proposed mechanism, we trained and tested eight encoder-decoder networks as summarized in Table \ref{tab:effect}. We compared our model to two conventional models: a plain encoder-decoder network \cite{Badrinarayanan2017:eke} (Figure \ref{fig:plainEncDec}) and a residual encoder-decoder \cite{fakhry2017residual:eke} (Figure \ref{fig:resEncDec}). Then, we ran ablation studies by removing IA-RCU blocks (A1), multiple decoders (A2), and both IA-RCU blocks and multiple decoders (A3). We ran all models with a single encoder-decoder network using single resolution input and with multiple encoder decoders using multiple resolution inputs. Finally, to compare with our fusion approach for multi-resolution inputs, we implemented two alternative fusion methods, (Figure \ref{fig:fuseTech}): Fusion-A, with a standard stack of convolutional blocks, and Fusion-B, with a spatial pyramid pooling method using atrous or dilated convolutions \cite{chen2016deeplab}. We used two resolutions in multi-resolution models but our network can be easily extended to many resolutions. 

\begin{figure}[t!]
  \centering
  \begin{tabular}{cc}
   \includegraphics[height=80px]{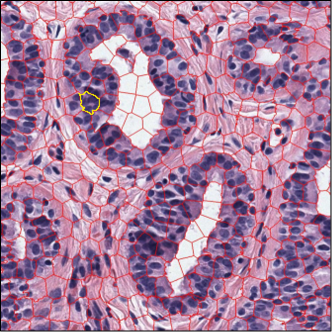}&\includegraphics[height=80px]{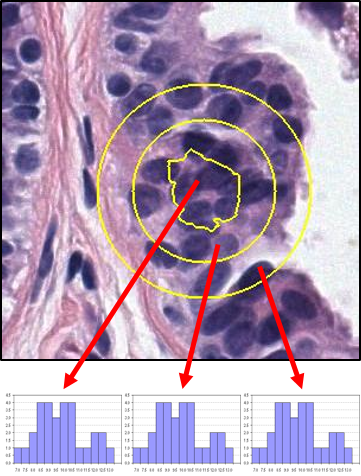} \\
   \small{(a) superpixel segmentation} & \small{(b) neighborhoods}
   \end{tabular}
  \caption{Initial superpixel segmentation and the circular neighborhoods used to increase the superpixel classification accuracy for supervised segmentation. Best viewed in color.}
  \label{fig:sp_context}
\end{figure}

\noindent \textbf{Superpixel and SVM-based Baseline:} For purpose of comparison, we also implemented a traditional feature-based segmentation method as a baseline. We refer to this method as SP-SVM. We used the SLIC algorithm \cite{Achanta2012} to segment H\&E images into superpixels of size 3,000 pixels. From each superpixel, we extracted color histograms on L*a*b* channels and LBP texture histograms \cite{He1990} on the H\&E channels. We used the color deconvolution algorithm \cite{Ruifrok2001} to separate the H\&E channels. A superpixel size of 3,000 pixels
was selected to have approximately one or two epithelial cells in one superpixel in order to capture detailed duct structures. To improve the classification, we included two circular neighborhoods around each superpixel in feature extraction. The color and texture histograms calculated from the superpixels and circular neighborhoods were concatenated to produce one feature vector for each superpixel. Figure \ref{fig:sp_context} illustrates the two circular neighborhoods from which the same features were extracted and appended to the superpixel feature vector.

\noindent \textbf{Training Details:} We split 58 images (regions of interest marked and annotated by the experts) into training (N=30) and test (N=28) sets. For the single-resolution networks, we cropped patches of size $256 \times 256$ with an overlap of 56 pixels at different WSI resolutions ($5\times$ and $10 \times$). For the multi-resolution networks, for each $256 \times 256$ patch, we created another patch by including a $64$-pixel border area (see Figure \ref{fig:proposedModel}). When necessary, we used symmetric padding to complete the patches. We obtained $5,312$ patches from the training set (N=30). To augment the data, we used standard augmentation strategies,  such as random rotations, horizontal flips, and cropping, resulting in a total of 25,992 patches. We used a 90:10 ratio for splitting these patches into training and validation sets. 

We trained all of our models end-to-end using stochastic gradient descent with a fixed learning rate of 0.0005, momentum of 0.9, weight decay of 0.0005, and a batch size of 10 on a single NVIDIA GTX-1080 GPU. We initialized encoder weight with ResNet-18 \cite{he2016deep:eke} trained on the ImageNet dataset \cite{krizhevsky2012imagenet}. We choose ResNet-18, because it: (1) is fast at inference, (2) requires less memory per image, and (3) learns less parameters while delivering accuracy similar to VGG \cite{simonyan2014very:eke} on the ImageNet. We initialized decoder weights as suggested in \cite{he2015delving}. We did not use dropout, following the practice of \cite{Ioffe2015BatchNA, he2016deep:eke}. We used an inverse class probability weighting scheme to deal with the class imbalance. Motivated by He \etal  \cite{he2016deep:eke}, we applied batch normalization \cite{Ioffe2015BatchNA} and ReLU \cite{he2015delving} operations after every convolution or deconvolution or atrous/dilated convolution operation, with the exception of RCU and IA-RCU blocks where second ReLU is performed after the element-wise sum operation.

For the superpixel and SVM-based baseline, we used the concatenated color and texture histograms to train an SVM that classifies super-pixels into eight tissue labels. To address the non-uniform distribution of the tissue labels and ROI size variation, we sampled 2,000 superpixels for each of the eight labels (if possible) from each image. We used the same training and test sets to evaluate the SP-SVM method. 

\begin{table*}[t!]
\centering
\resizebox{\textwidth}{!}{
\small
\begin{tabular}{rccc||c|ccc||c|ccc}
\toprule[1pt]\midrule[0.3pt]
& {Dense} & {Multi-} & {IA-} & \multicolumn{4}{c||}{\textbf{Single resolution}} &  \multicolumn{4}{c}{\textbf{Multiple resolution}} \\
\cline{5-12}
& {Conn.} & {Dec.} & {RCU} & \textbf{\# Params} & \textbf{F1} & \textbf{mIOU}  & \textbf{PA} & \textbf{\# Params} &\textbf{F1} & \textbf{mIOU}  & \textbf{PA}\\ 
\cmidrule{1-12}
Plain Enc-Dec \cite{Badrinarayanan2017:eke} &   &   &   & 12.80 M & 0.507 & 0.376  & 0.575 & 25.61 M & 0.513 & 0.381  & 0.593 \\
Residual Enc-Dec \cite{fakhry2017residual:eke} &   &   &   & 12.80 M & 0.510 & 0.381 & 0.586 & 25.61 M & 0.517 & 0.386  & 0.597 \\
\textbf{Our Model}  &\checkmark &\checkmark & \checkmark & 13.00 M & 0.554 & 0.418 & 0.642 & 26.03 M & \textbf{0.588} & \textbf{0.442} & \textbf{0.700} \\ 
\cmidrule{1-12}
A1 & \checkmark & \checkmark &  & 12.93 M & 0.517 & 0.385 & 0.608 & 25.85 M & 0.529 & 0.390 & 0.631 \\
A2 & \checkmark &   & \checkmark & 12.99 M & 0.517 & 0.387  & 0.601 & 25.98 M & 0.540 & 0.407 & 0.633 \\
A3 & \checkmark &   &  & 12.92 M & 0.519 & 0.390 & 0.607 & 25.84 M & 0.524 & 0.392  & 0.611 \\
\cmidrule{1-12}
Ours + Fusion-A &\checkmark &\checkmark & \checkmark & NA & NA & NA & NA & 26.03 M & 0.535 & 0.402 & 0.631 \\ 
Ours + Fusion-B &\checkmark &\checkmark & \checkmark & NA & NA & NA & NA & 26.00 M & 0.554 & 0.419 & 0.658 \\
\midrule[0.3pt]\midrule[0.3pt]
SP-SVM & \multicolumn{3}{c||}{NA} & NA & 0.365 & 0.258 & 0.485 & NA & NA & NA & NA\\
\midrule[0.3pt]\bottomrule[1pt]
\end{tabular}
}
\caption{Quantitative comparison of different methods on the Breast  Biopsy dataset.}
\label{tab:effect}
\end{table*}

\subsection{Segmentation Results}
We evaluated our results using three metrics commonly used for semantic segmentation \cite{Chen2016:eke, shelhamer2017fully:eke, Badrinarayanan2017:eke}: (1) F1-score (F1), (2) mean region Intersection over Union (mIOU), and (3) global pixel accuracy (PA). Table \ref{tab:effect} summarizes the performance of different encoder-decoder models and feature-based baseline. The impact of each of our modifications along with a comparison with the feature-based segmentation method are discussed below. 

\noindent \textbf{Residual vs Dense Connections:} The residual encoder-decoder has a $0.5\%$ higher pixel accuracy (PA) than the plain encoder-decoder, and our model with dense connections (A3) has a $2\%$ higher PA than plain encoder-decoder under both single and multiple resolution settings. On an average, dense connections improve the accuracy (across different metrics) by at least $1\%$ without significantly increasing the number of parameters of the network.

\begin{figure}[t!]
\setlength{\belowcaptionskip}{-2mm}
\centering
\resizebox{0.85\columnwidth}{!}{
\setlength\tabcolsep{2pt}
\begin{tabular}{ccccc}
\midrule[0.3pt]
\multicolumn{1}{c}{RGB} & \multicolumn{2}{||c||}{RCU} & \multicolumn{2}{|c}{IA-RCU} \\
\midrule
\includegraphics[width=1.5cm, height=1.5cm]{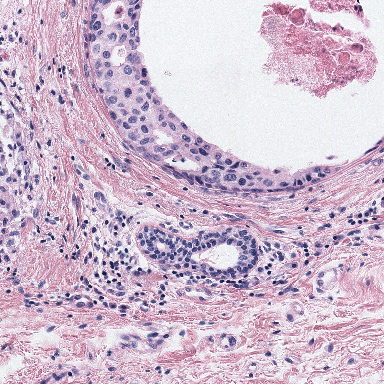} &  \includegraphics[width=1.5cm, height=1.5cm]{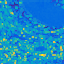} & \includegraphics[width=1.5cm, height=1.5cm]{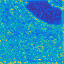} &  \includegraphics[width=1.5cm, height=1.5cm]{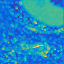} & \includegraphics[width=1.5cm, height=1.5cm]{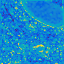} \\
\includegraphics[width=1.5cm, height=1.5cm]{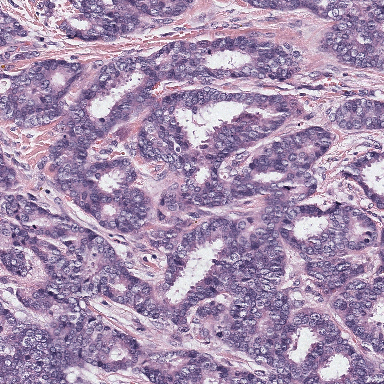} &  \includegraphics[width=1.5cm, height=1.5cm]{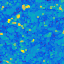} & \includegraphics[width=1.5cm, height=1.5cm]{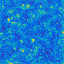} &  \includegraphics[width=1.5cm, height=1.5cm]{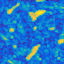} & \includegraphics[width=1.5cm, height=1.5cm]{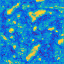} \\
\midrule
Size $\rightarrow$ & $64 \times 64$ & $32\times 32$ & $64 \times 64$ & $32\times 32$
\end{tabular}
}
\resizebox{0.08\columnwidth}{!}{
\begin{tabular}{c}
\includegraphics[width=\columnwidth]{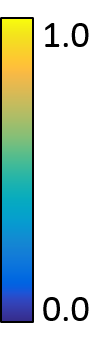}
\end{tabular}
}
\caption{Visualization of activation maps of different encoding blocks at different spatial resolutions. IA-RCU compensates the loss of spatial information due to down-sampling operations and helps in learning features that are relevant with respect to input. For visualization, we have scaled the activation maps to the same spatial dimensions. Best viewed in color.}
\label{fig:impactIAF}
\end{figure}

\noindent \textbf{RCU vs IA-RCU:} Replacing the IA-RCU with conventional RCUs (A1) in our model reduces accuracy (both F1 and PA) by about $4\%$ under single resolution and $7\%$ under multiple resolutions. Furthermore, A2 with IA-RCU has $2\%$ higher accuracy than A3 with RCUs under multiple resolution setting. Figure \ref{fig:impactIAF} visualizes the activation maps of different encoding blocks at different spatial resolutions in which RCUs lose information about small structures in lower spatial dimensions, while the IA-RCUs help in retaining this information.

\noindent \textbf{Single vs Multiple Decoders:} Replacing multiple decoders with a single decoder in A2 reduces the pixel accuracy of our full model by $4\%$ with single resolution and $7\%$ with multiple resolutions. Furthermore, A1 has $2\%$ higher pixel accuracy than A3 under multiple resolution setting. The pixel accuracy does not change from A3 to A1 under the single resolution setting. 

\begin{figure}[t!]
\setlength{\belowcaptionskip}{-2mm}
\centering
\resizebox{0.85\columnwidth}{!}{
\begin{tabular}{cccc}
\includegraphics[width=0.2\columnwidth]{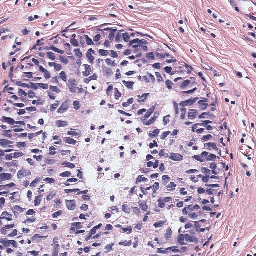} & \includegraphics[width=0.2\columnwidth]{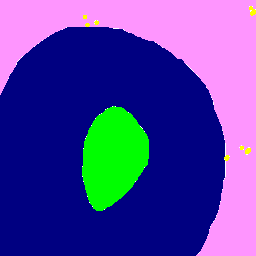} & \includegraphics[width=0.2\columnwidth]{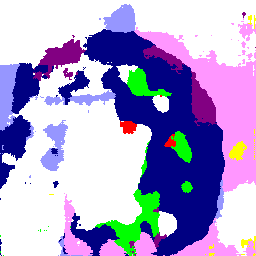} & \includegraphics[width=0.2\columnwidth]{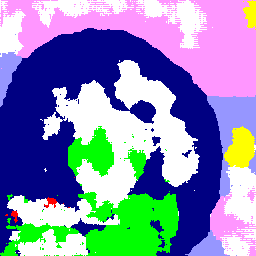} \\
\includegraphics[width=0.2\columnwidth]{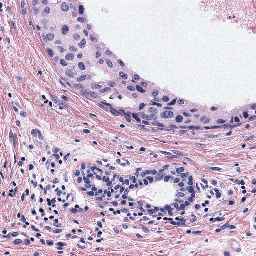} & \includegraphics[width=0.2\columnwidth]{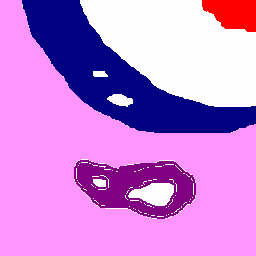} & \includegraphics[width=0.2\columnwidth]{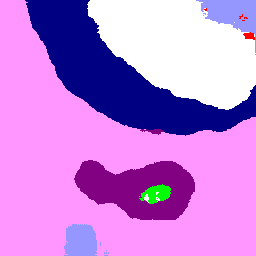} & \includegraphics[width=0.2\columnwidth]{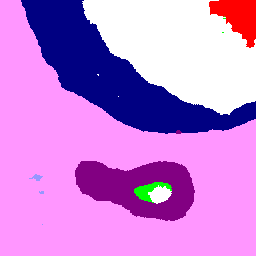} \\
\small{RGB} & \small{Ground} & \small{Plain Model} & \small{Plain Model} \\
\small{Patch} & \small{Truth} & \small{(single)} & \small{(multi)} \\ 
\end{tabular}
}
\caption{Patch-wise predictions of Plain Encoder-Decoder network with single and multiple resolution input. Multi-resolution input helps in improving the predictions, especially at the patch borders. Best viewed in color.}
\label{fig:visSingVsMult}
\end{figure}

\begin{figure}[t!]
\setlength{\belowcaptionskip}{-2mm}
\resizebox{\columnwidth}{!}{
\begin{tabular}{ccccc}
     \includegraphics[width=0.19\columnwidth]{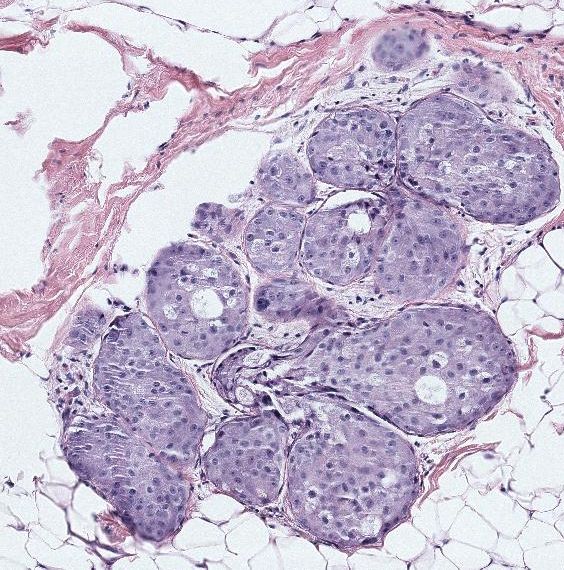}&
     \includegraphics[width=0.19\columnwidth]{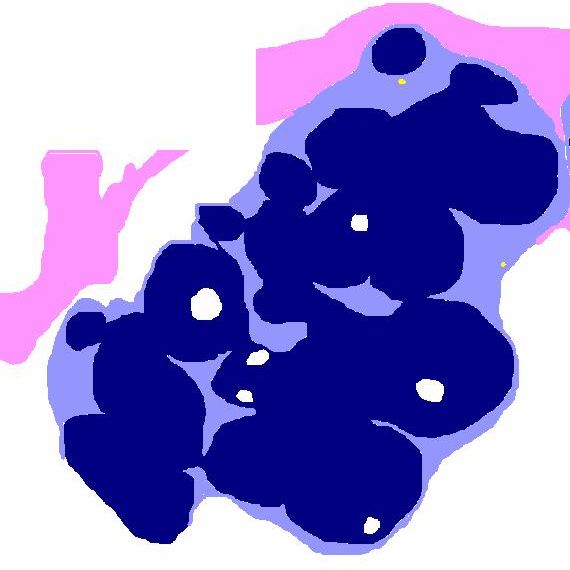}&
     \includegraphics[width=0.19\columnwidth]{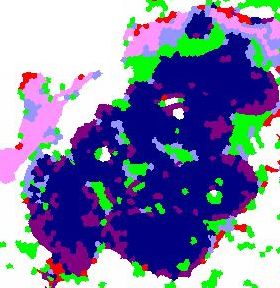}&
     \includegraphics[width=0.19\columnwidth]{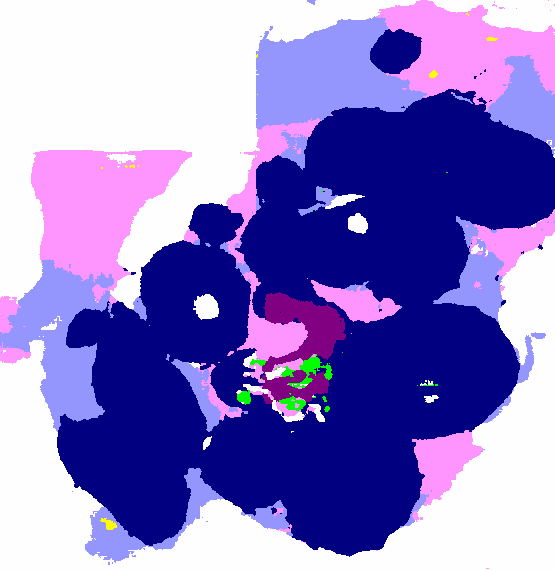}&
     \includegraphics[width=0.19\columnwidth]{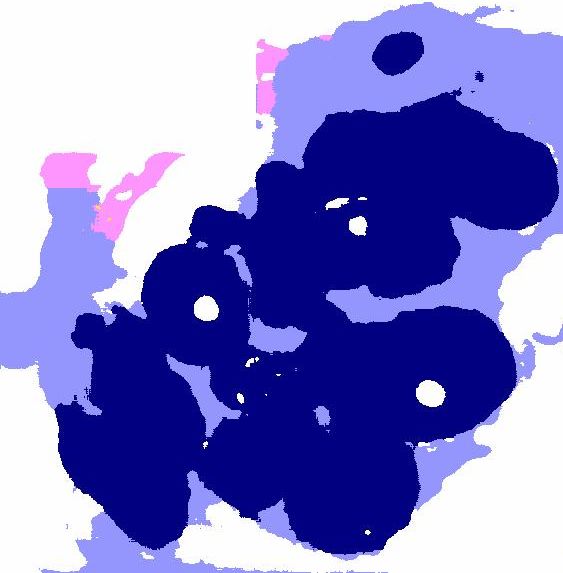}\\
     \includegraphics[width=0.19\columnwidth]{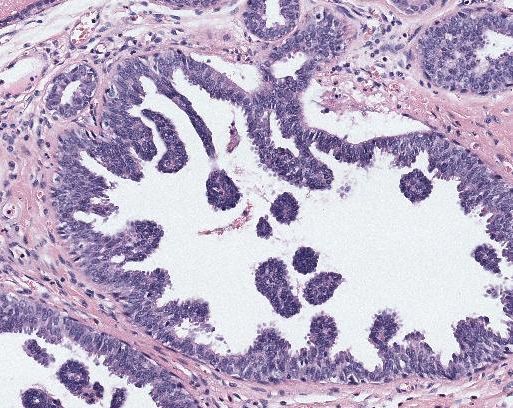}&
     \includegraphics[width=0.19\columnwidth]{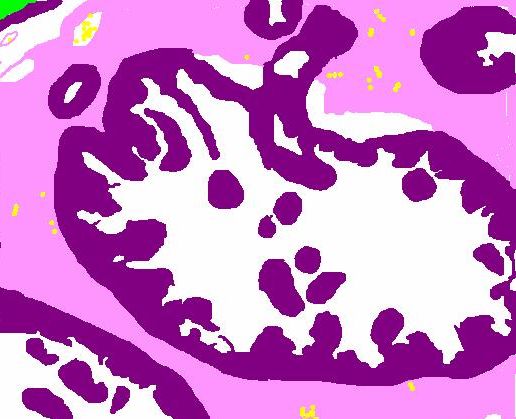}&
     \includegraphics[width=0.19\columnwidth]{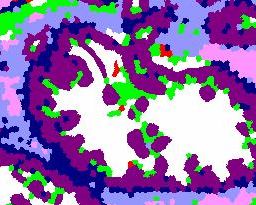}&
     \includegraphics[width=0.19\columnwidth]{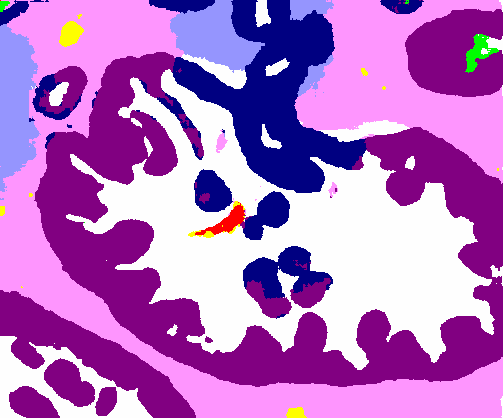}&
     \includegraphics[width=0.19\columnwidth]{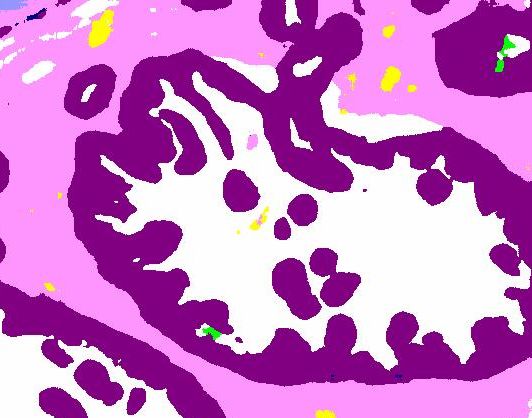}\\
     RGB & Ground & SVM  & Our Model & Our Model \\
     & Truth & & (single) & (multi)
\end{tabular}
}
\caption{ROI-wise predictions: first row depicts an invasive case while the second row depicts a benign case. Best viewed in color.}
\label{fig:segWSI}
\end{figure}

\noindent \textbf{Single vs Multiple Resolutions:} For all models, multi-resolution inputs improve the performance up to 6\% in  pixel  accuracy. All metrics increase from single resolution to multi-resolution for all models. Although the improvement in accuracy is small, multi-resolution input leads to better segmentation results (see Figure \ref{fig:visSingVsMult} and Figure \ref{fig:segWSI}). 


\noindent \textbf{Different Fusion Methods:} The overall F1-score of our model with our fusion scheme (Figure \ref{fig:proposedModel}) is about $6\%$ and $4\%$ higher than Fusion-A and Fusion-B  (Figure \ref{fig:fuseTech}), respectively. 

\begin{figure}[b!]
\centering
\includegraphics[width=0.95\columnwidth]{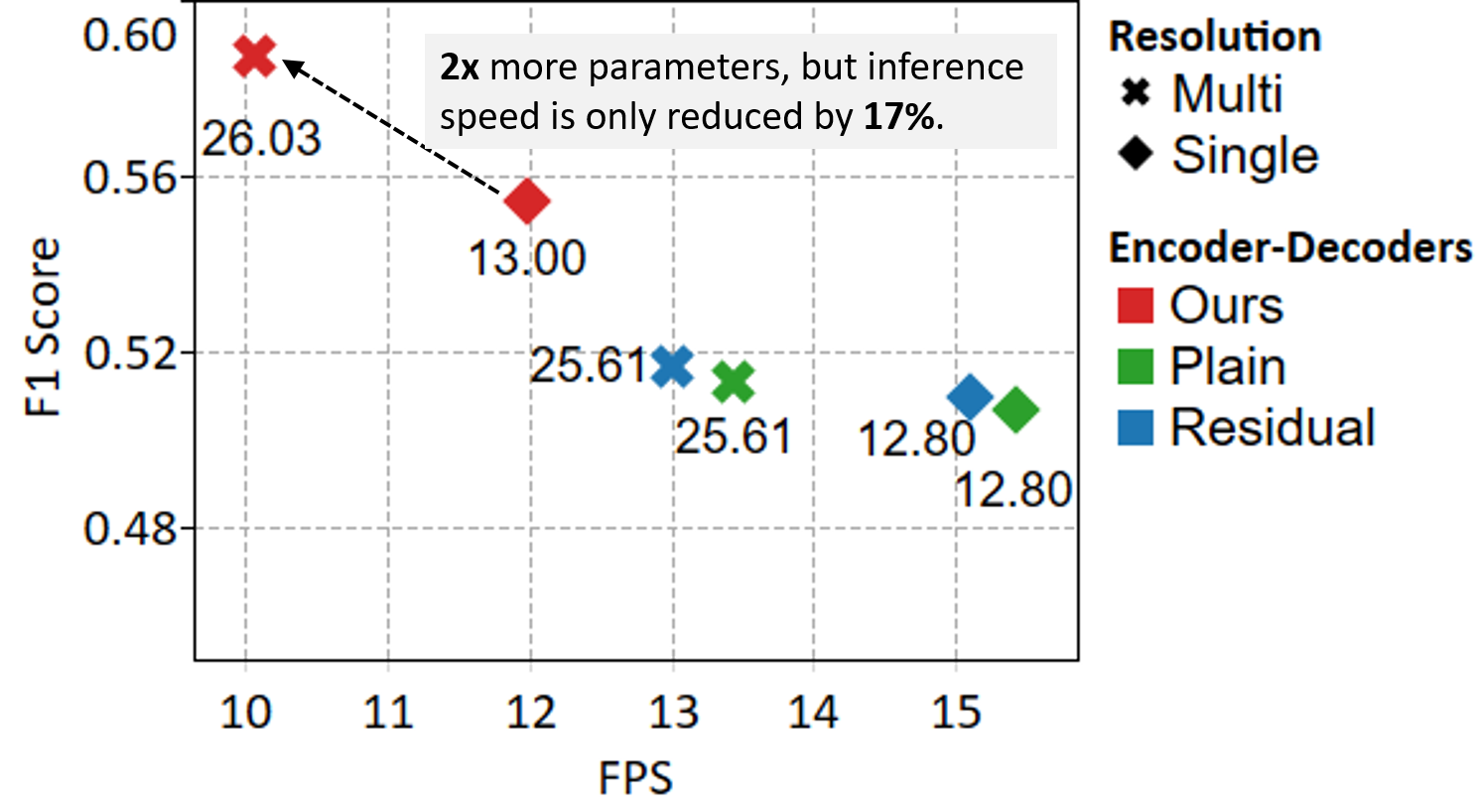}
\caption{Impact on inference time and number of parameters learned at different resolutions. Number of parameters are in million and are listed next to the corresponding data point. Inference time is measured on NVIDIA GTX-1080 GPU and is an average across 3 trials for 20 samples of size $384\times384$. Here, FPS refers to frames (or patches) processed per second. Best viewed in color.}
\label{fig:timevsparam}
\end{figure}

\noindent \textbf{Inference Time and Number of Parameters:} The impact on inference time and number of parameters learned by both single and multi-resolution networks is reported in Figure \ref{fig:timevsparam}. Multi-resolution network utilize the hardware resources efficiently by executing multiple encoder-decoder networks simultaneously and therefore, the impact on inference time is not drastic. The multi-resolution networks are merely $0.2\times$ slower than the single resolution network while learning almost $2\times$ more parameters.

\noindent \textbf{Comparison with Feature-Based Baseline:} Since the SP-SVM method used only single resolution images, we compared it to our model's performance with single resolution input. Our model outperformed the SP-SVM method across all metrics.

\subsection{Diagnostic Classification}
Semantic segmentation provides a powerful abstraction for diagnostic classification. We designed a set of experiments to show the descriptive power of the tissue label segmentation in automated diagnosis. To this end, we used the full set of ROIs (N=428) to predict the consensus diagnosis assigned by the expert panel. We trained and tested two types of classifiers, an SVM and a multi-layer perceptron (MLP), for four classification tasks: (1) 4-class (benign vs. atypia vs. DCIS vs. invasive); (2) invasive vs. non-invasive (benign, atypia and DCIS); (3) benign vs. non-benign (atypia and DCIS); and (4) atypia vs. DCIS. The last three tasks were designed to imitate the diagnostic decision making process of pathologists while the first one is the naive approach. 

We applied our model with single and multiple-resolutions and SP-SVM-based baseline to all the images in our dataset (N=428) to get tissue label segmentations. For diagnostic features, we calculated the frequency and co-occurrence histograms of superpixel tissue labels, using the majority pixel label for the CNN approach that labels pixels.
We trained SVMs and MLPs for the four classification tasks in a 10-fold cross-validation setting and repeated the experiments 10 times. During training, we subsampled the data to have a uniform distribution of diagnostic classes.

\begin{figure}[t!]
\centering
\includegraphics[width=\columnwidth]{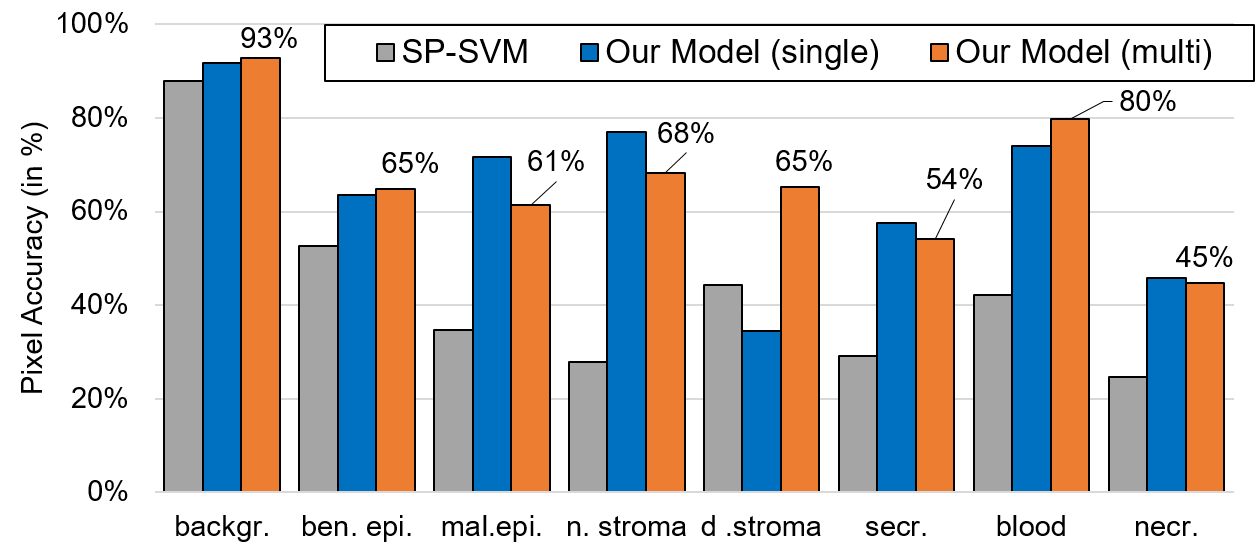}
\caption{Segmentation accuracy for different labels. Best viewed in color.}
\label{fig:seg_bars}
\end{figure}

\noindent \textbf{Results:} The accuracies for four diagnostic classification tasks are given in Table \ref{tab:diag_acc}. The features calculated from segmentation masks produced by our model outperforms the SP-SVM method with both classifiers, with the exception of classification of benign cases with SVM. In particular, multi-resolution input improves the segmentation of desmoplastic stroma label significantly (Figure \ref{fig:seg_bars}), which is easily identifiable in lower-resolutions and an important tissue type for diagnosing breast cancer \cite{mao2013stromal}. Incorporating input from larger surrounding tissue helps the model identify tumor-associated desmoplastic stroma, in turn, it improves the classification of invasive cases (90.7\% with the multi-resolution model and SVM classifier). 

We note that the separation between classes using only the distribution of tissue labels is clear from our results, suggesting that tissue label images have high descriptive power.


\begin{table*}[t!]
\centering
\resizebox{\textwidth}{!}{
\small
\begin{tabular} {r||cc|cc|cc||cc|cc|cc}
\toprule[1pt]\midrule[0.3pt]
& \multicolumn{6}{c||}{Diagnostic Classifier: \textbf{SVM}} &\multicolumn{6}{c}{Diagnostic Classifier: \textbf{MLP}}\\
\cline{2-13}
& \multicolumn{2}{c|}{\textbf{SP-SVM}} & \multicolumn{2}{c|}{\textbf{Our Model (single)}} & \multicolumn{2}{c||}{\textbf{Our Model (multi)}} & \multicolumn{2}{c|}{\textbf{SP-SVM}} & \multicolumn{2}{c|}{\textbf{Our Model (single)}} & \multicolumn{2}{c}{\textbf{Our Model (multi)}}\\
\cline{2-13}
& all & no & all & no & all & no & all & no & all & no & all & no \\
& labels & stroma & labels & stroma & labels & stroma & labels & stroma & labels & stroma & labels & stroma \\
\hline
\textbf{4-class} & 35.5\% & 32.1\% & 44.5\% & 36.3\% & \textbf{45.9\%} & 36.3\% & 45.0\% & 38.6\% & \textbf{54.5\%} & 46.4\% & 54.2\% & 45.2\% \\
\hline
\textbf{invasive} & 64.7\% & 44.6\% & 78.4\% & 58.4\% & \textbf{90.7\%} & 63.4\% & 69.0\% &57.8 \% & 69.0\% & 64.1\% & \textbf{76.0\%} & 68.7\% \\
\textbf{benign}  & 55.0\% & \textbf{67.7\%} & 44.7\% & 65.3\% & 40.0\% & 61.0\% & 61.1\% & 60.3\% & \textbf{66.5\%} & 66.2\% & 65.8\% & 64.2\% \\
\textbf{atypia-DCIS} & 66.34\% & 59.2\% & 84.69\% & \textbf{85.1\%} & 84.07\% & 82.8\% & 74.28\% & 68.5\% & 85.03\% & \textbf{87.7\%} & 82.07\% & 81.3\% \\
\midrule[0.3pt]\bottomrule[1pt]
\end{tabular}
}
\caption{Diagnostic classification accuracies for different classification methods}
\label{tab:diag_acc}
\end{table*}

\section{Discussion}
Diagnostic classification with the full range of breast diagnoses is a difficult problem. In a previous study, a group of pathologists interpreted the same digital slides of breast biopsies \cite{Elmore2015} and achieved accuracies of 70\%, 98\%, 81\% and 80\% for the tasks of 4-class, invasive vs. (benign-atypia-DCIS), (atypia-DCIS) vs. benign, and DCIS vs. atypia respectively. Semantic segmentation provides a powerful abstraction so that simple features with diagnostic classifiers, like SVM and multi-layer perceptron, perform well in comparison to pathologists. 

Multi-resolution input increases the context of the model and improve the segmentation of the labels pathologists identify in lower resolutions; such as desmoplastic stroma. Furthermore, our fusion block outperforms the alternative fusion blocks, most likely due to its high effective receptive field. The effective receptive field of our block (Figure \ref{fig:proposedModel}) is $65 \times 65$ while the effective receptive fields of the fusion blocks in Figure \ref{fig:f1} and \ref{fig:f2} are $7\times 7$ and $37\times37$. In addition to quantitative evaluation, our model results in smoother borders for the segmented regions while the SP-SVM method is limited to color similarity for initial segmentation and has much smaller context than networks. 

An automated diagnosis system should operate on whole slide images. Since the whole-slide-level annotations were not available on our data, we validated our model on regions of interest that were identified, diagnosed, and annotated by the experts. Our method can easily be applied to WSIs for segmentation or can be used in combination with a region of interest identifier for classification \cite{mercan2014roi}. Our future work involves developing a system for simultaneous ROI localization, segmentation, and diagnostic classification on WSIs. 

\section{Conclusions}
Our model outperforms traditional encoder-decoders and the SP-SVM-baseline both qualitatively and quantitatively (see Figure \ref{fig:segWSI}). It also improves the F1-score and mIOU of conventional networks by at least $7\%$ and the global pixel accuracy by $11\%$ for multiple resolution settings. This improvement is mainly due to the long-range direct connections that are established between input and output either using identity or projection mappings. These long-range connections helps in back-propagating the information directly to the input paths efficiently and therefore, improves the flow of information inside the network and eases the optimization.

We showed that our semantic segmentation provides powerful features for diagnosis. With hand-crafted or learned features for diagnosis, our model is promising for a computer-aided system for breast cancer diagnosis. 
Though we study breast biopsy images in this paper, our system can be easily extended to other types of cancer.

\section*{Acknowledgements}
Research reported in this publication was supported by the National Cancer Institute awards R01 CA172343, R01 CA140560, and KO5 CA104699. The content is solely the responsibility of the authors and does not necessarily represent the views of the National Cancer Institute or the National Institutes of Health. We thank Ventana Medical Systems, Inc. (Tucson, AZ, USA), a member of the Roche Group, for the use of iScan Coreo Au™ whole slide imaging system, and HD View SL for the source code used to build our digital viewer. For a full description o f HD View SL, please see \url{http://hdviewsl.codeplex.com/}. 

The authors thank Hannaneh Hajishirzi, Srinivasan Iyer, Michael Driscoll, Safiye Celik, and Sandy Kaplan for their thorough and helpful comments.

\appendix

\section{More Segmentation Results}
\begin{figure*}[t!]
\centering
\begin{tabular}{cccccc}
RGB & Ground Truth & Our Model & RGB & Ground Truth & Our Model \\
\includegraphics[width=70px, height=70px]{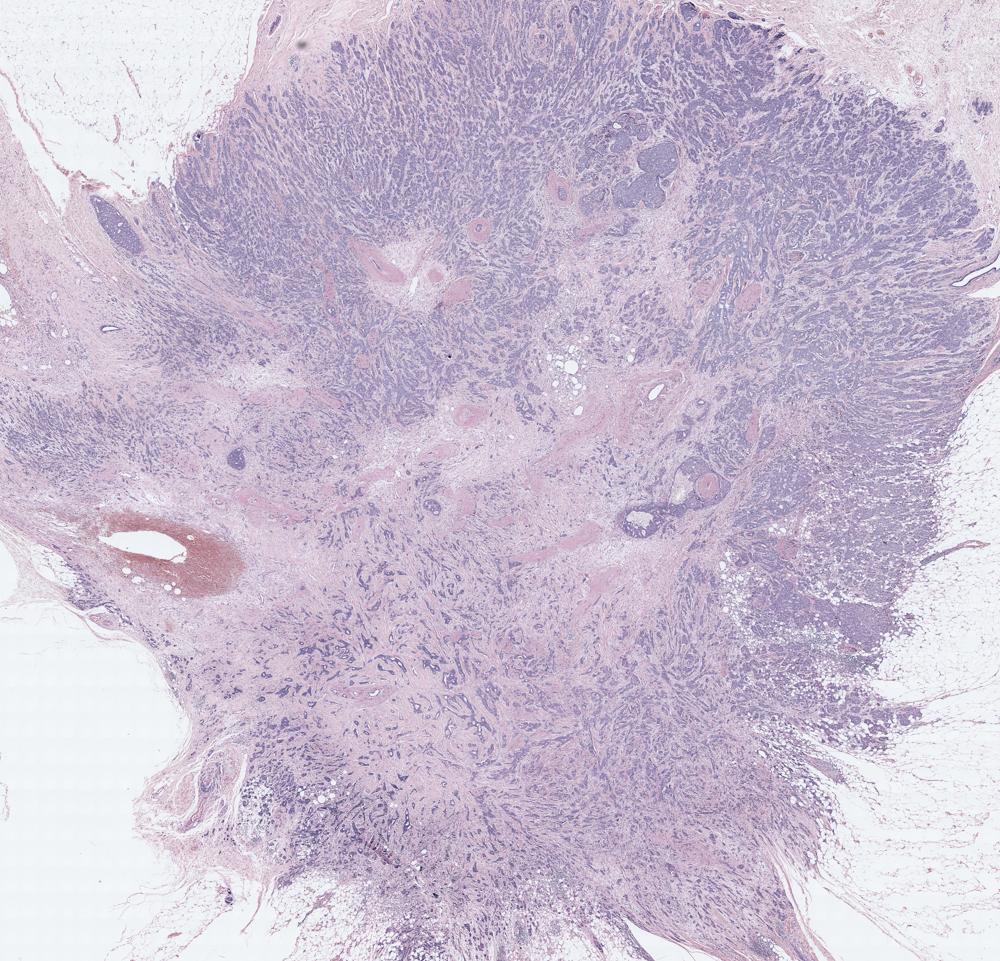}& \includegraphics[width=70px, height=70px]{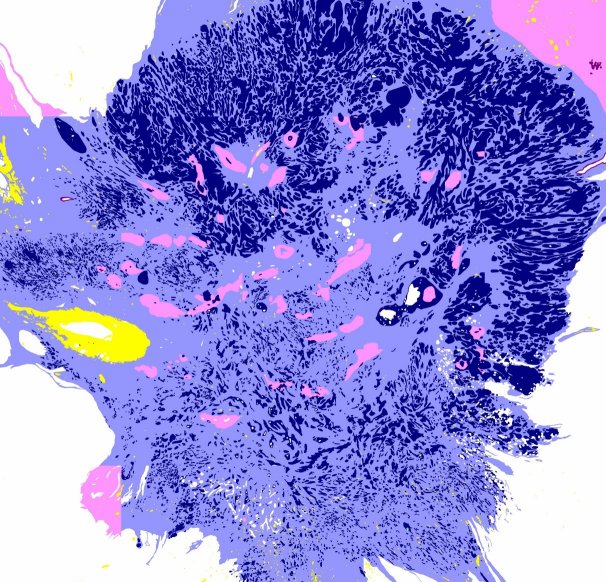}& \includegraphics[width=70px, height=70px]{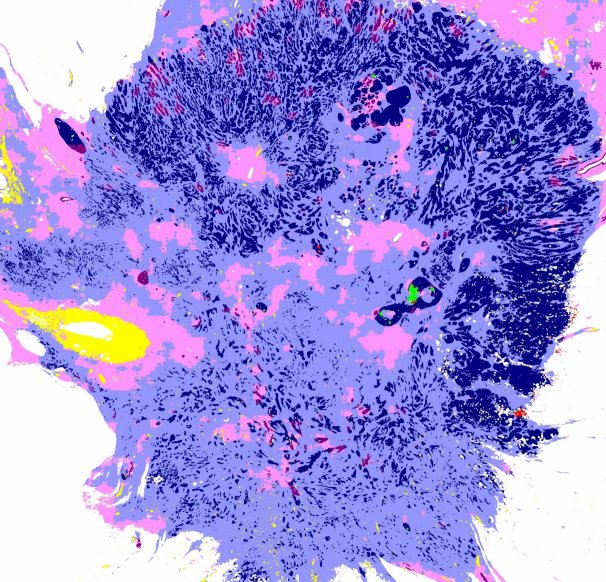} & \includegraphics[width=70px, height=70px]{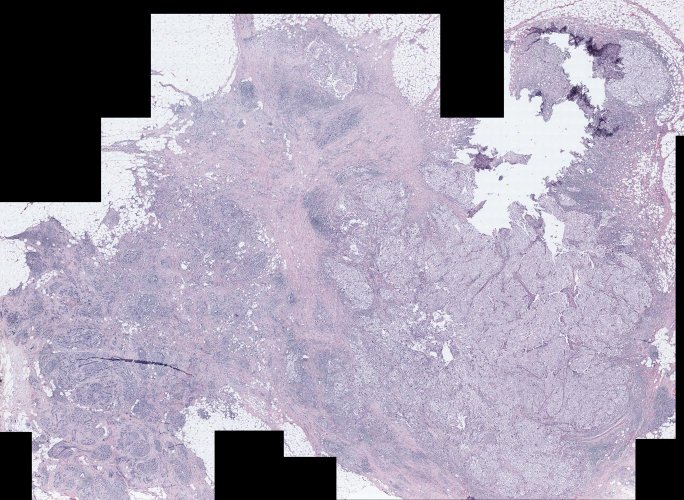} & \includegraphics[width=70px, height=70px]{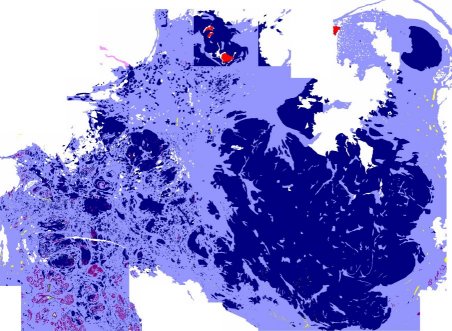} & \includegraphics[width=70px, height=70px]{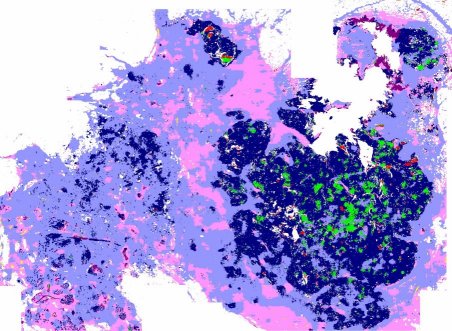} \\
\includegraphics[width=70px, height=70px]{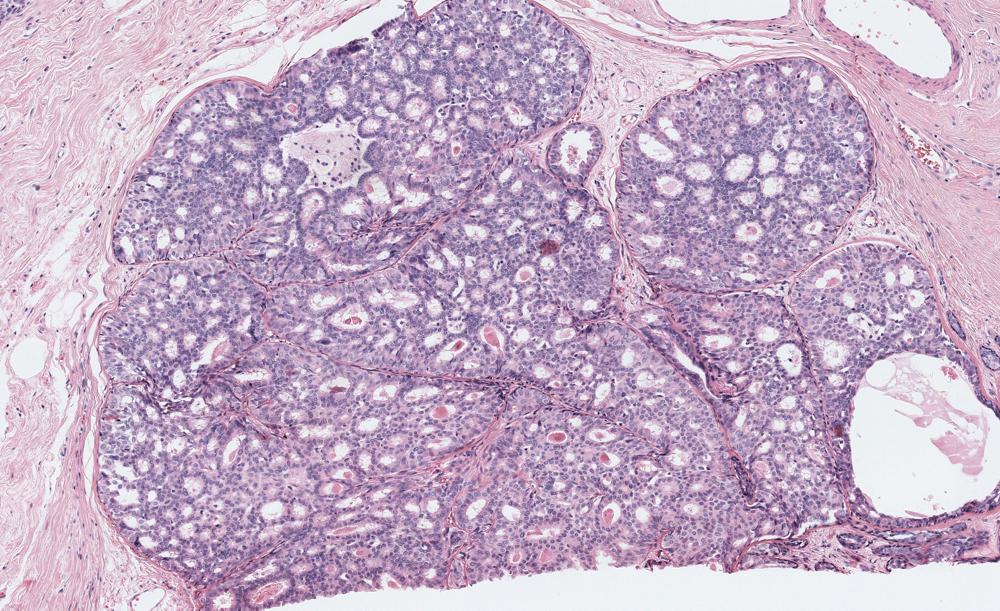}& \includegraphics[width=70px, height=70px]{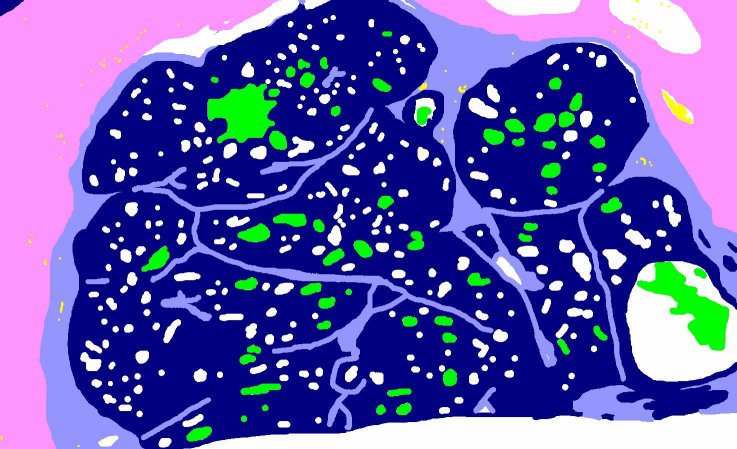}& \includegraphics[width=70px, height=70px]{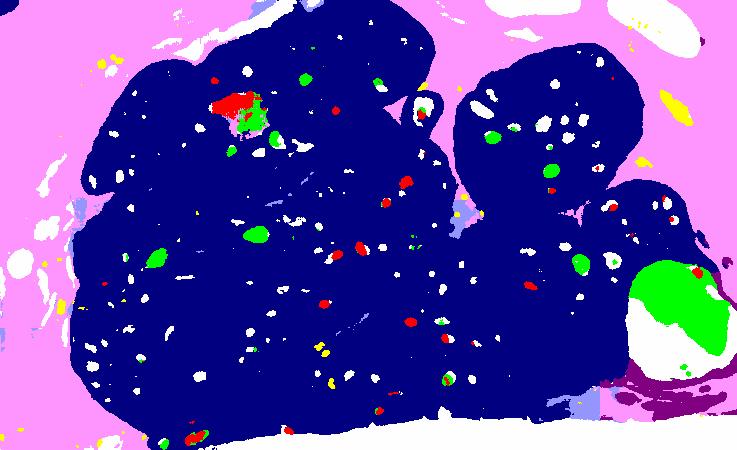} & \includegraphics[width=70px, height=70px]{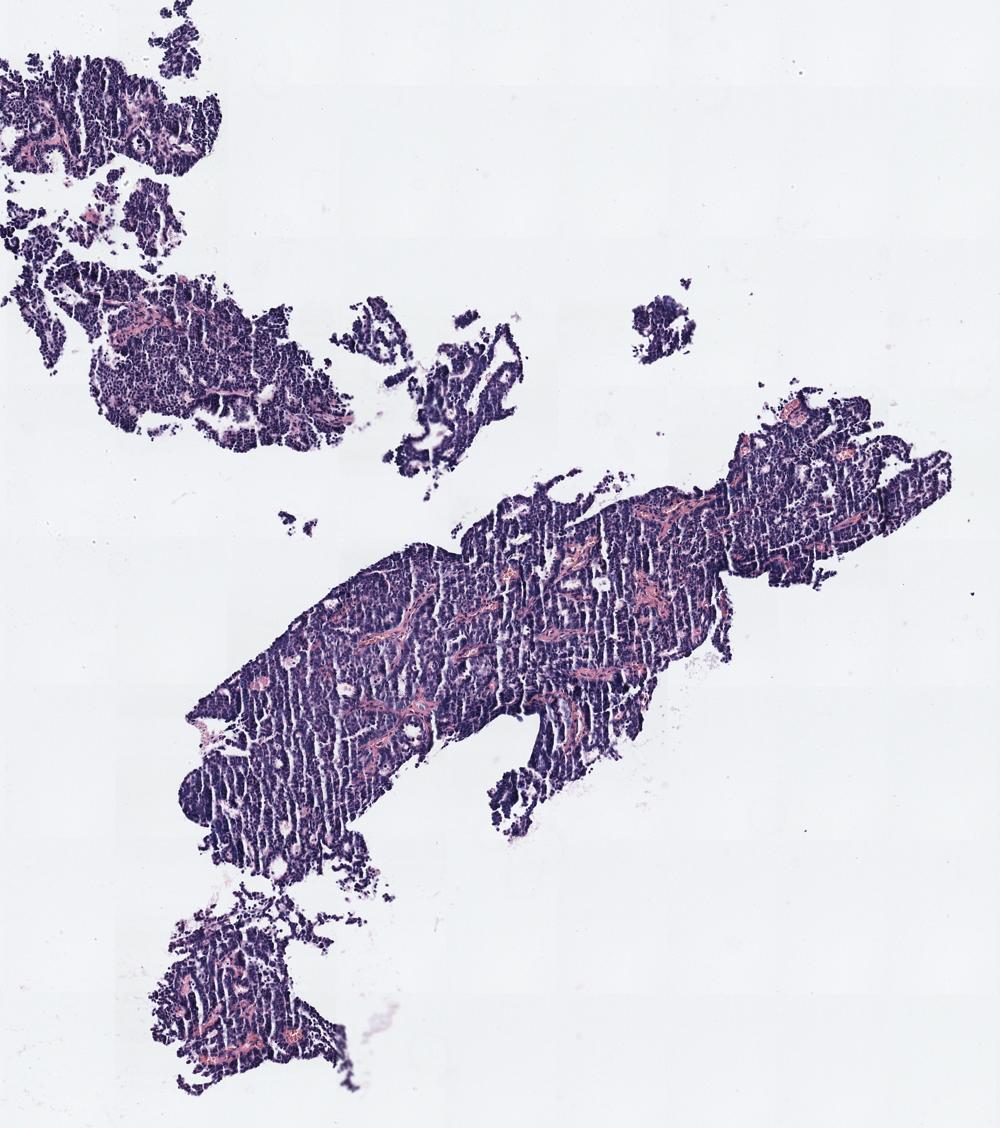} & \includegraphics[width=70px, height=70px]{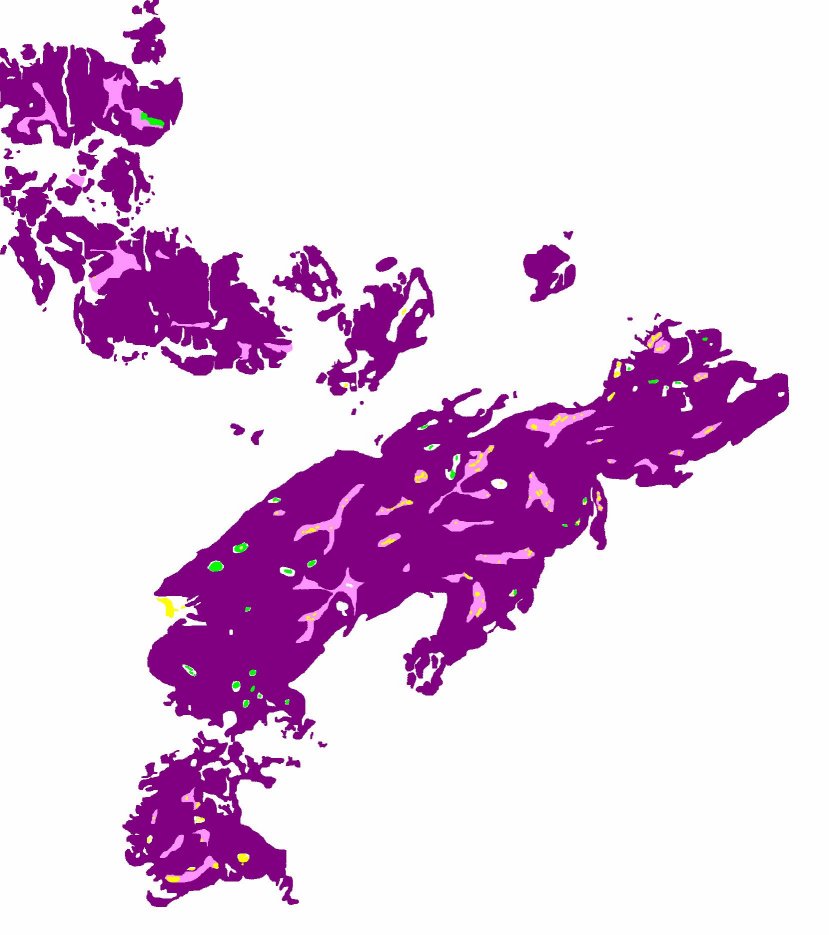} & \includegraphics[width=70px, height=70px]{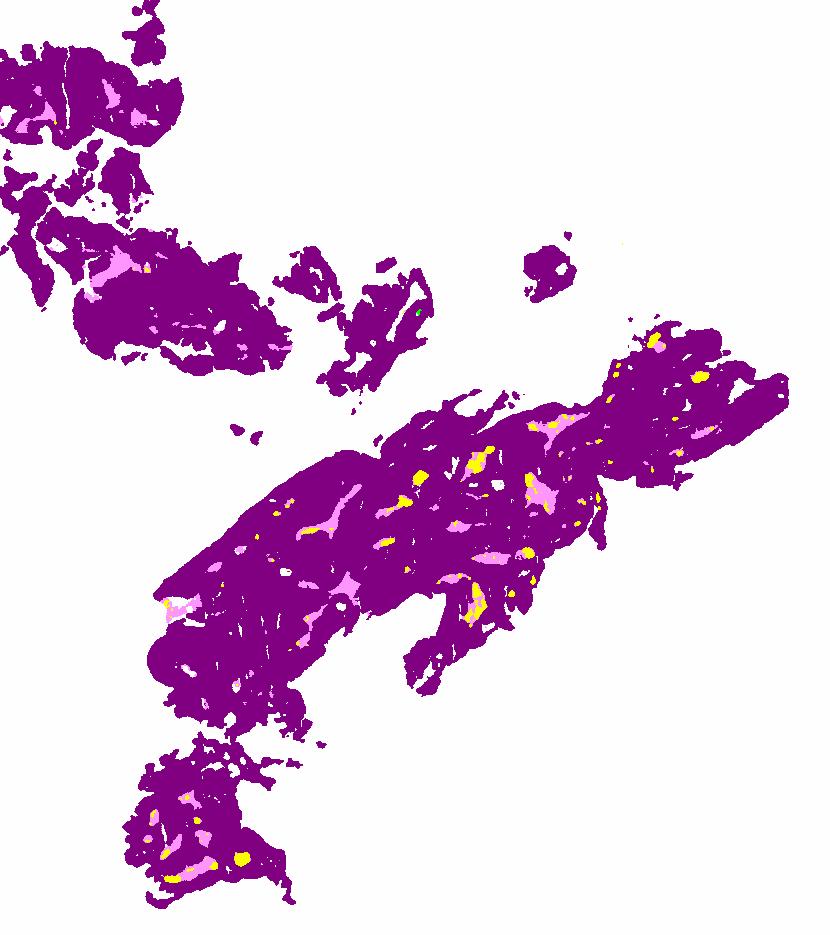} \\
\includegraphics[width=70px, height=70px]{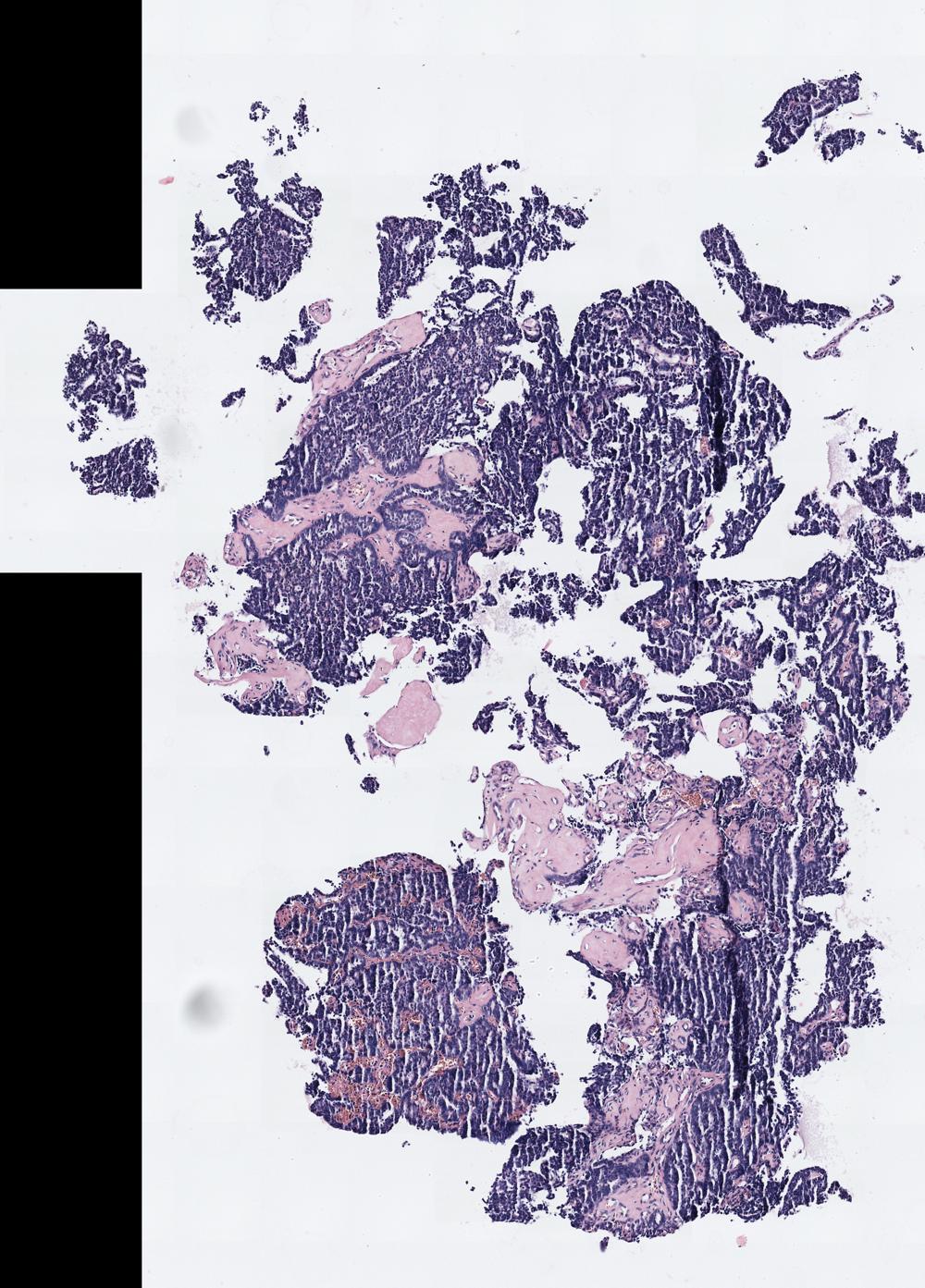}& \includegraphics[width=70px, height=70px]{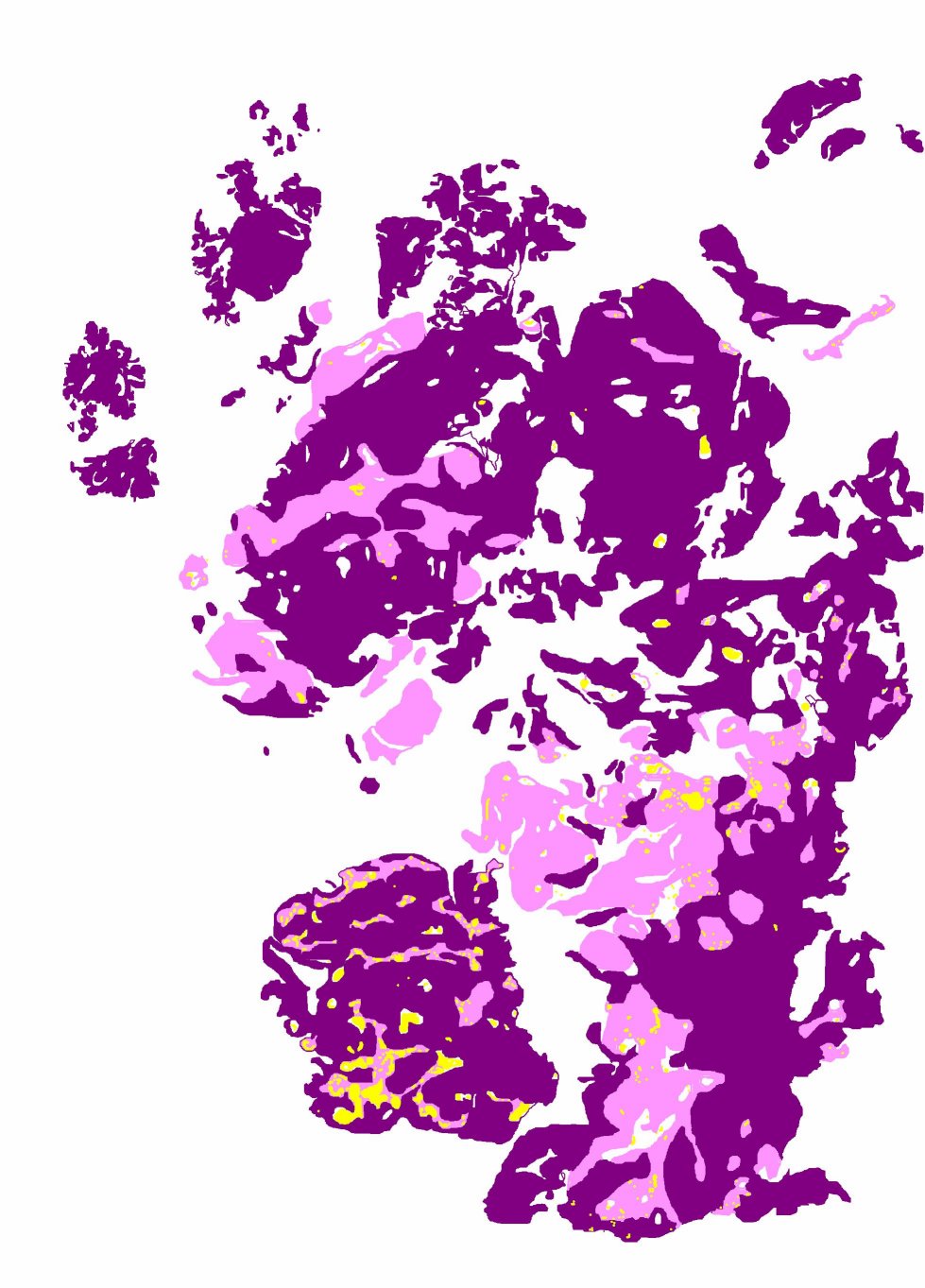}& \includegraphics[width=70px, height=70px]{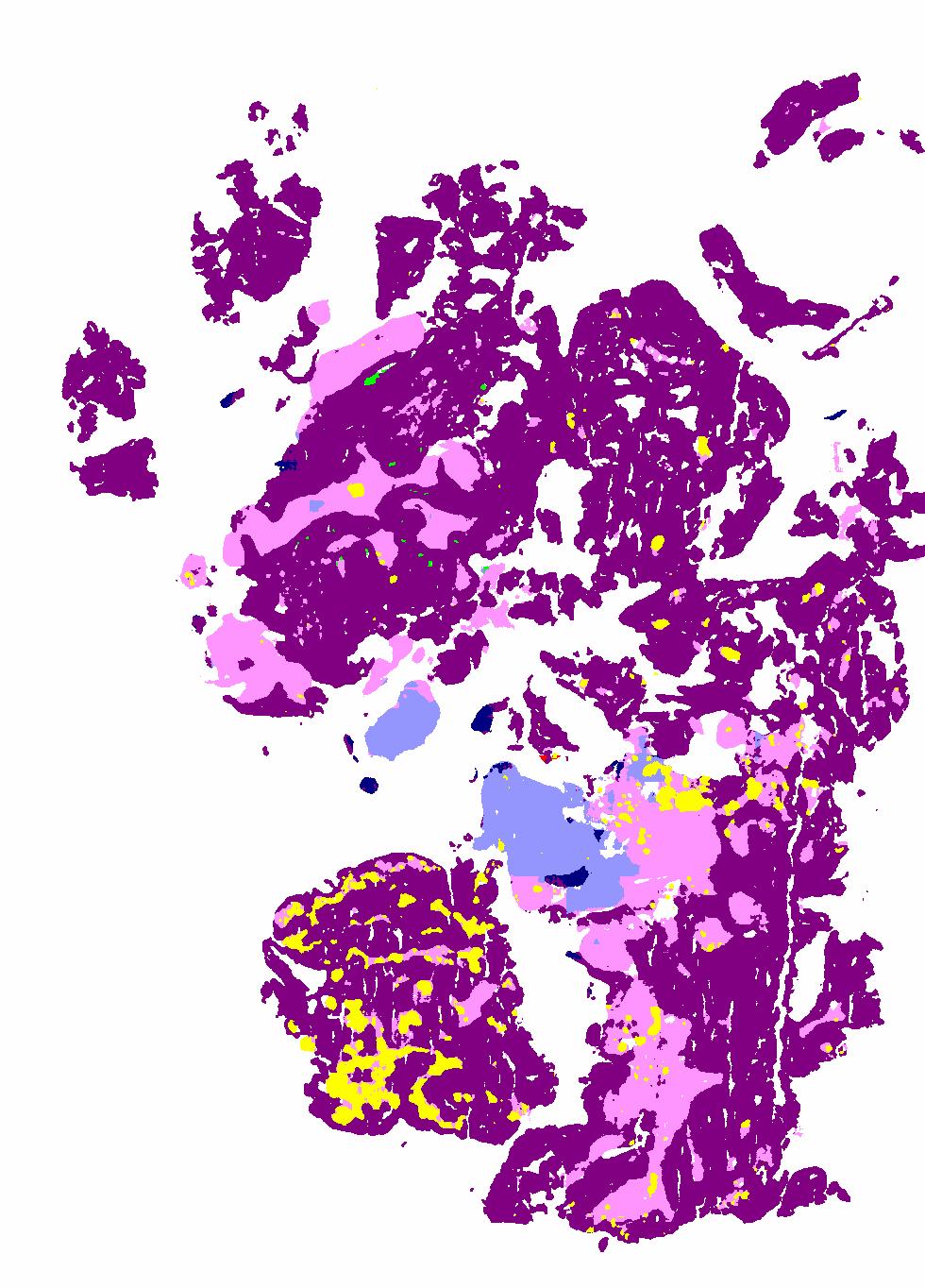} & \includegraphics[width=70px, height=70px]{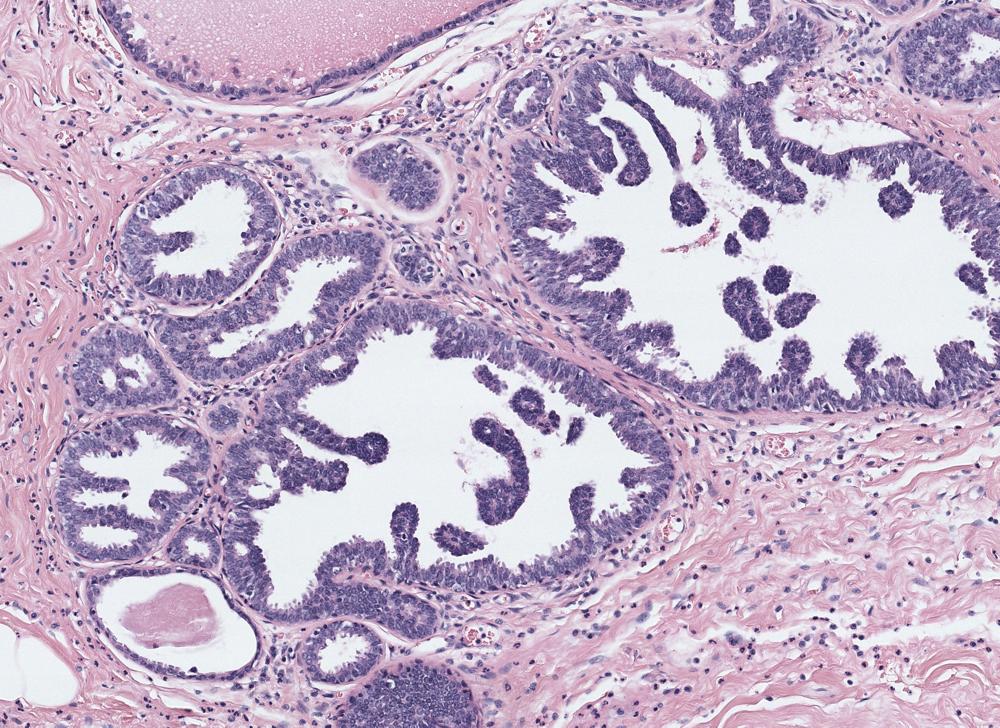} & \includegraphics[width=70px, height=70px]{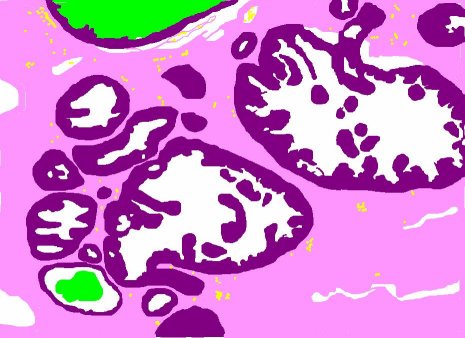} & \includegraphics[width=70px, height=70px]{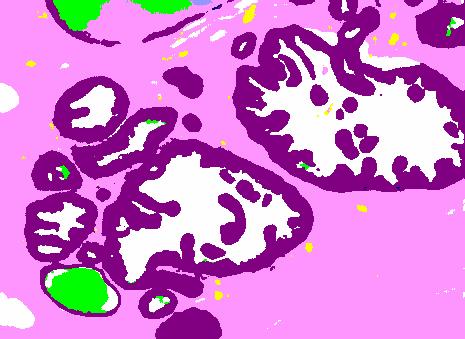} \\
\includegraphics[width=70px, height=70px]{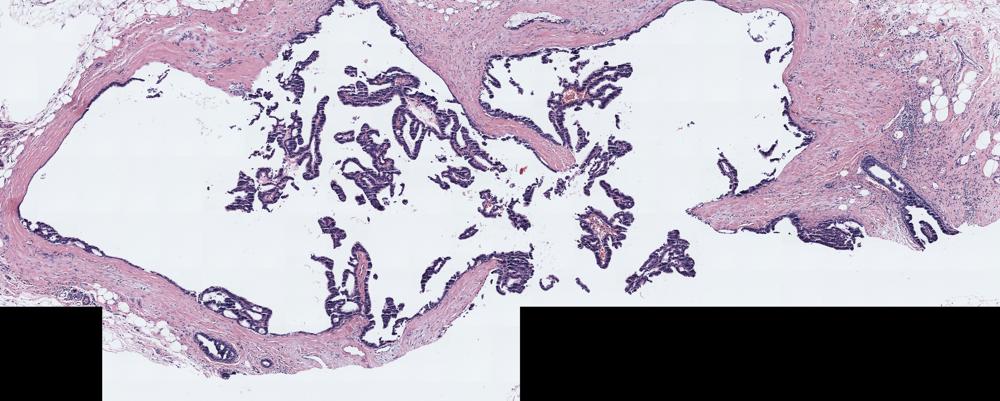}& \includegraphics[width=70px, height=70px]{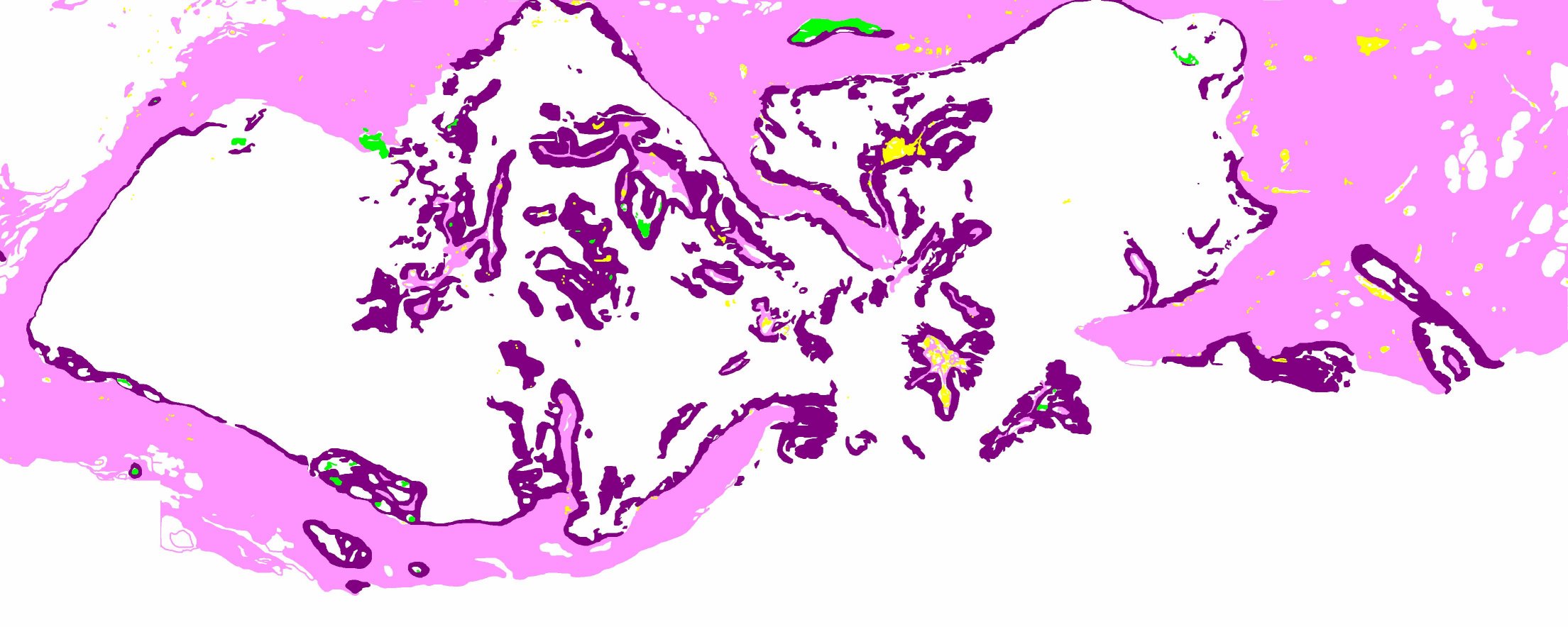}& \includegraphics[width=70px, height=70px]{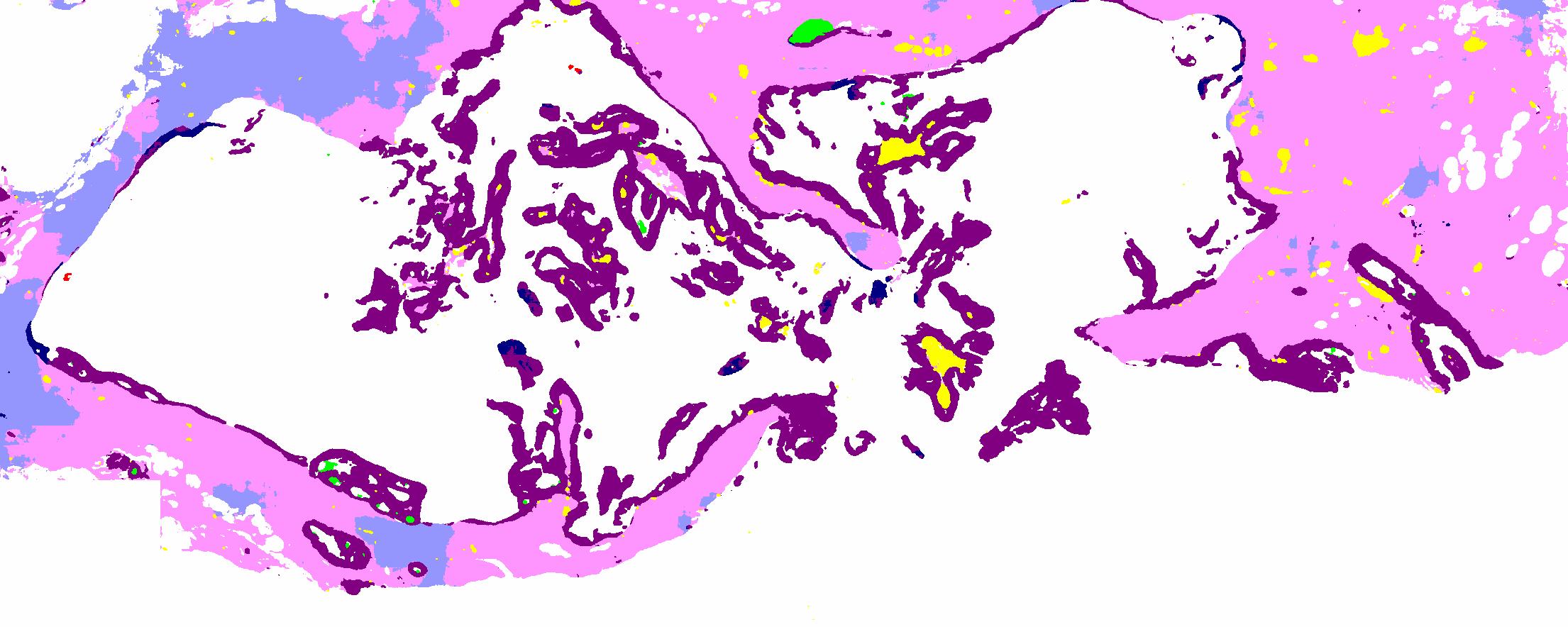} & \includegraphics[width=70px, height=70px]{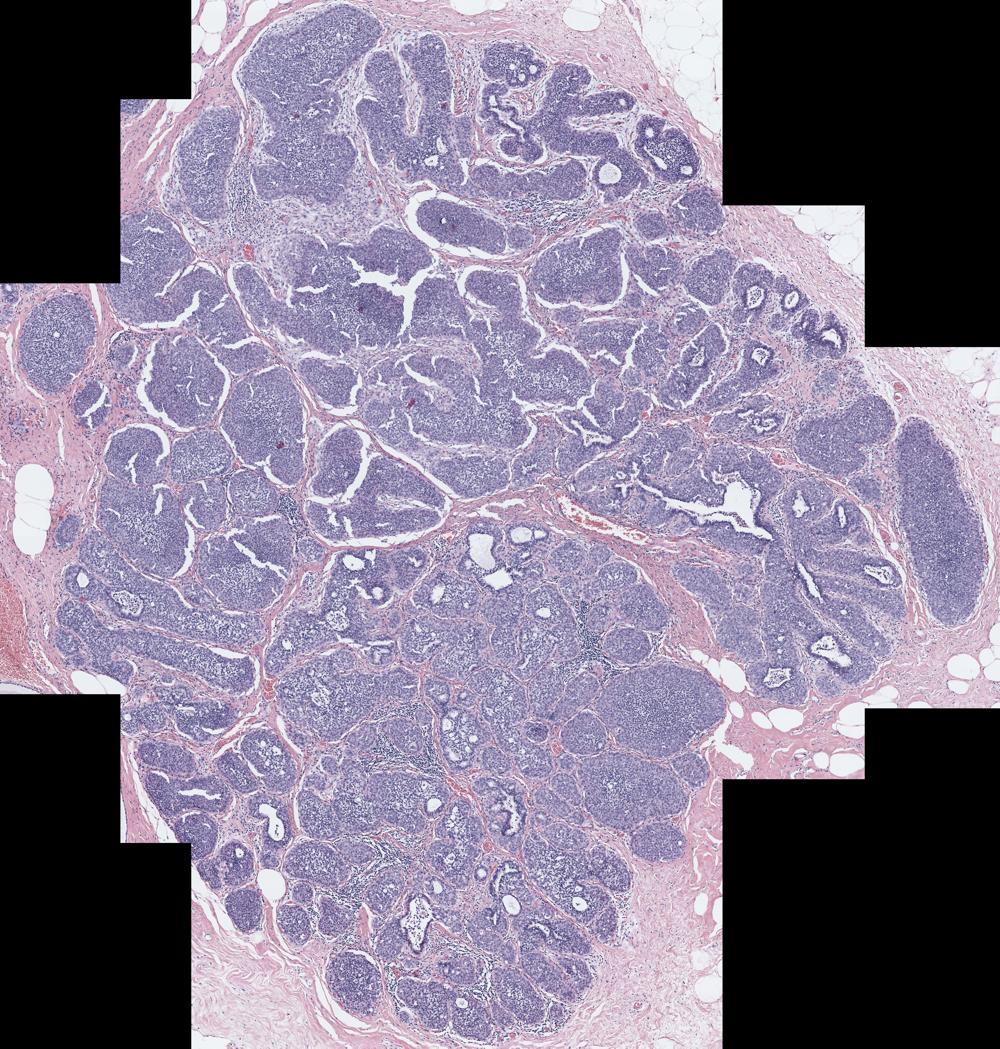} & \includegraphics[width=70px, height=70px]{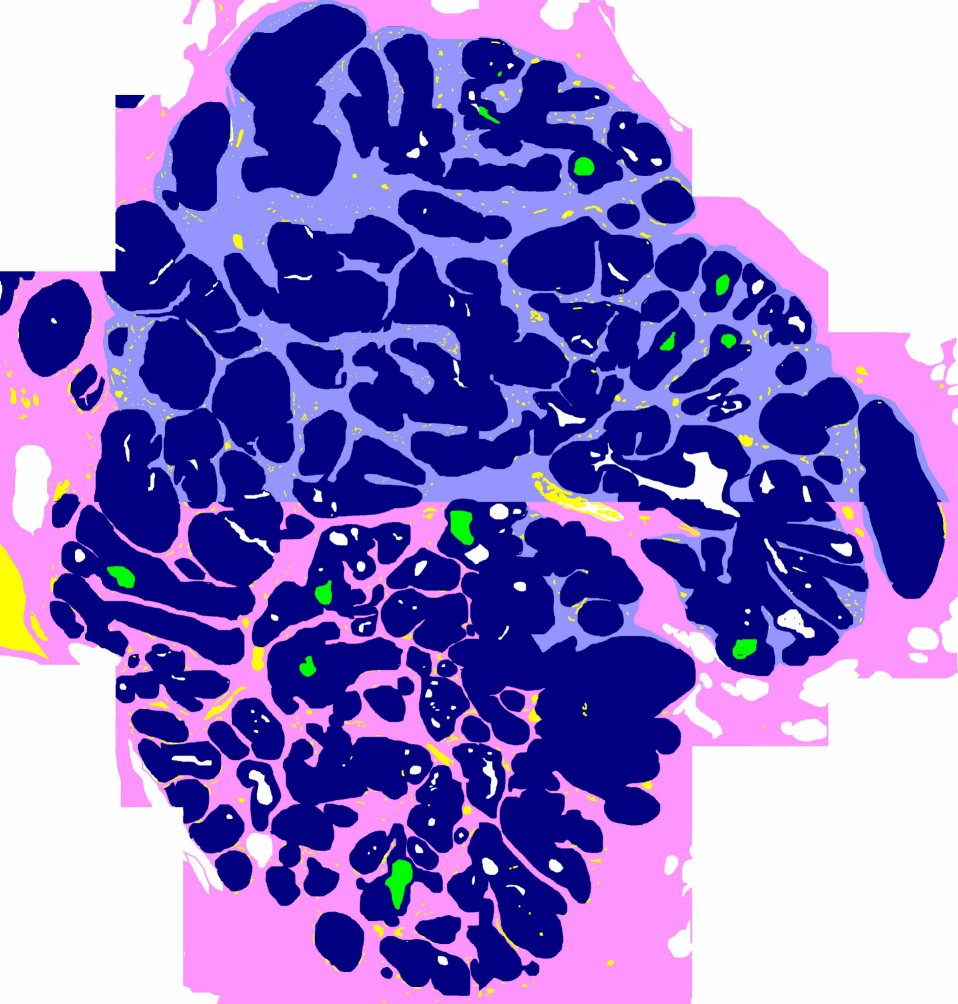} & \includegraphics[width=70px, height=70px]{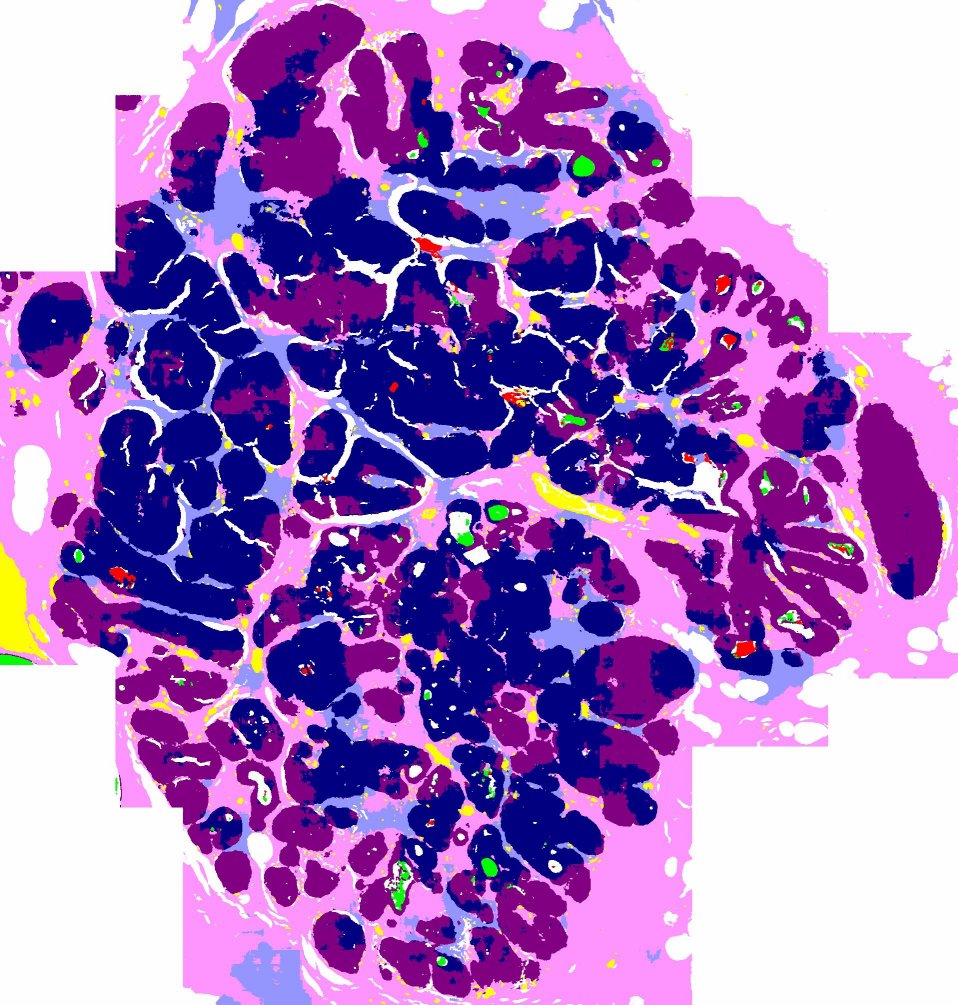} \\
\includegraphics[width=70px, height=70px]{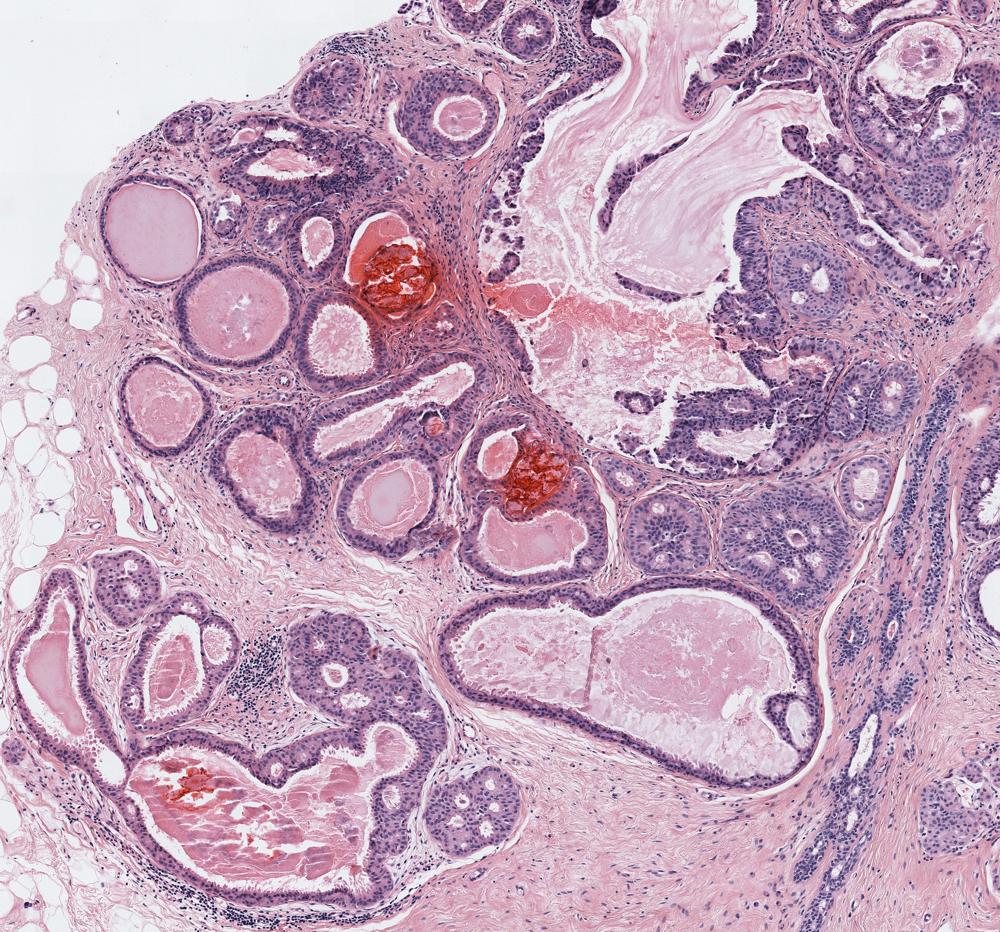}& \includegraphics[width=70px, height=70px]{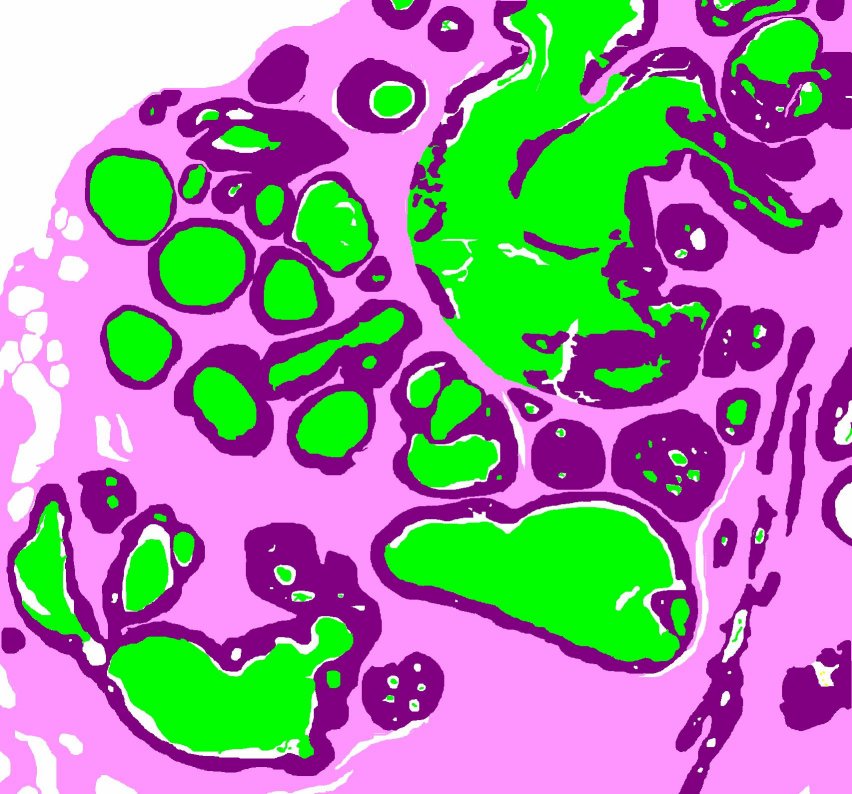}& \includegraphics[width=70px, height=70px]{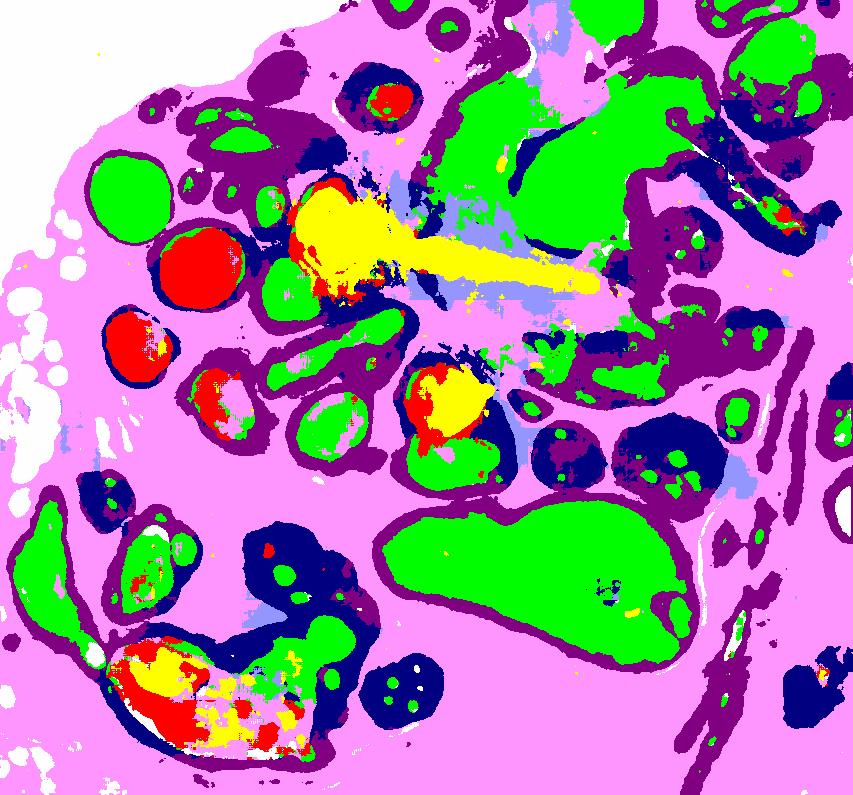} & \includegraphics[width=70px, height=70px]{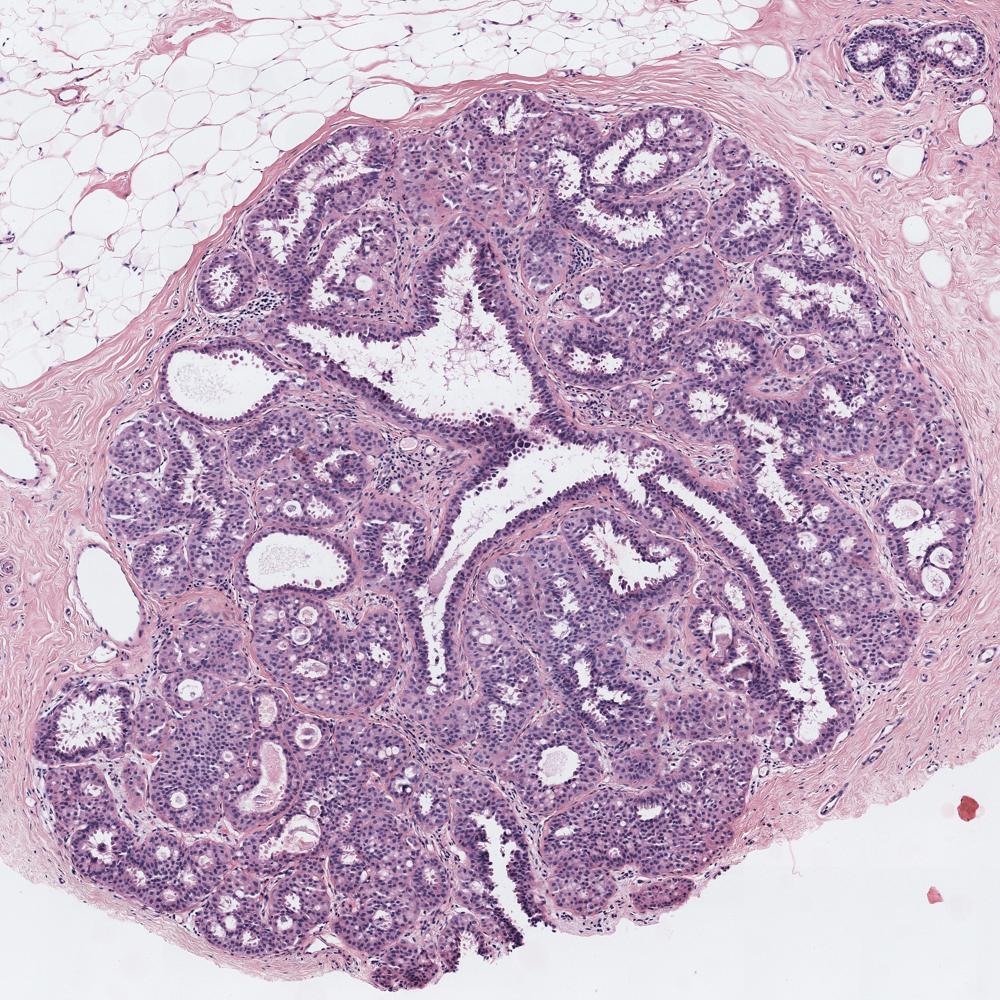} & \includegraphics[width=70px, height=70px]{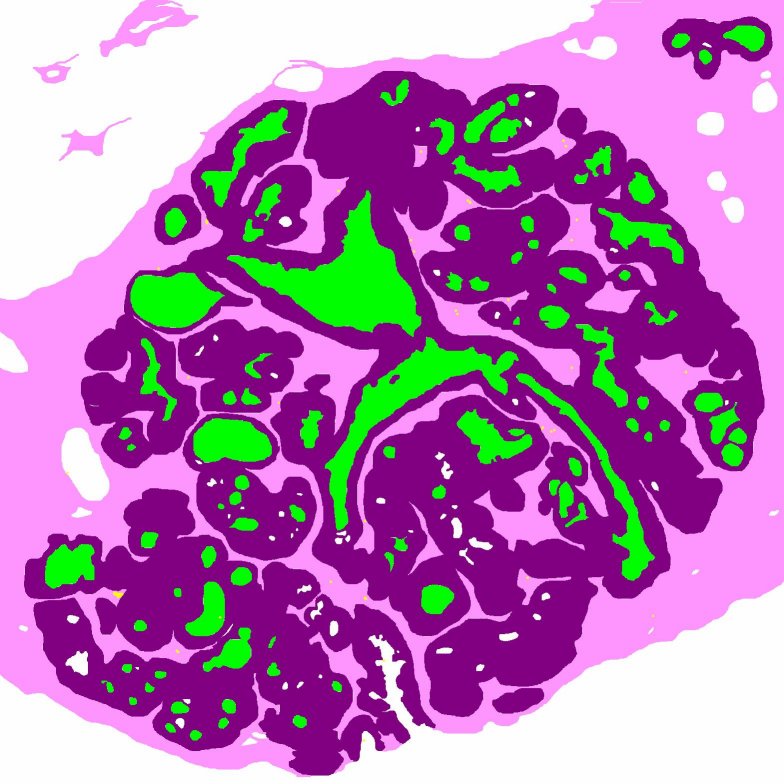} & \includegraphics[width=70px, height=70px]{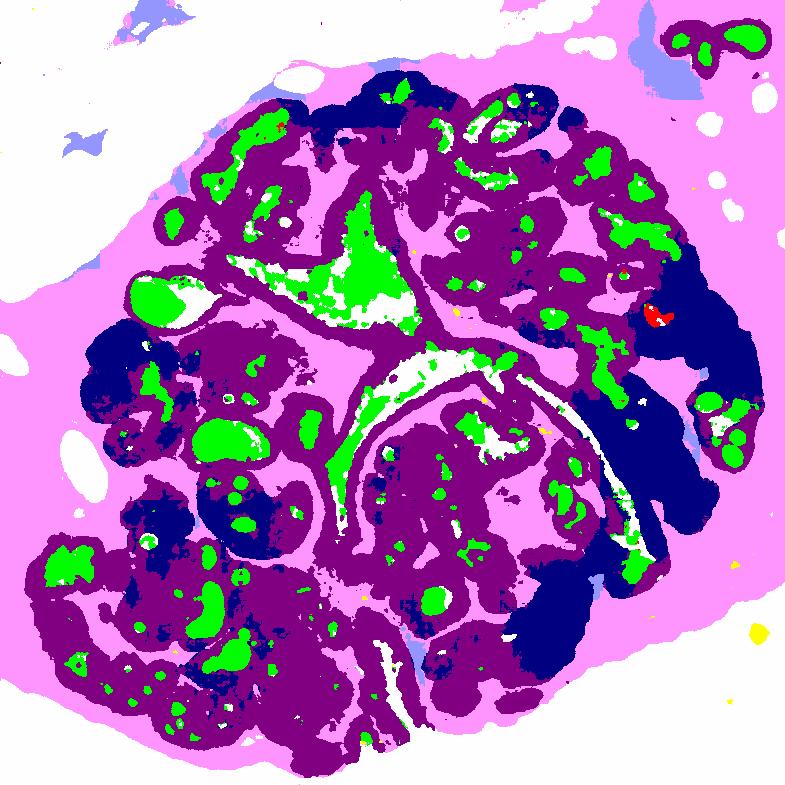} \\
\multicolumn{6}{c}{
\includegraphics[width=1.5\columnwidth]{images/label_legend.pdf}}
\end{tabular}
\caption{ROI-wise predictions of our method with multiple inputs on the validation set.}
\label{fig:roiPredMore}
\end{figure*}

Figure \ref{fig:roiPredMore} shows ROI-wise predictions of our method. From Figure \ref{fig:roiPredMore}, we can see that our method is promising and is able to address the challenges that WSIs present i.e. our method is able to segment the WSIs into different types of tissue labels irrespective of the tissue or WSI size.

{\small
\bibliographystyle{ieee}
\bibliography{medBib}
}

\end{document}